\documentclass{article} 
\usepackage{iclr2027_conference,times}

\usepackage{amsmath,amsfonts,bm}

\def\eqref#1{equation~\ref{#1}}

\def\1{\bm{1}}

\DeclareMathAlphabet{\mathsfit}{\encodingdefault}{\sfdefault}{m}{sl}
\SetMathAlphabet{\mathsfit}{bold}{\encodingdefault}{\sfdefault}{bx}{n}

\usepackage{graphicx}
\usepackage{booktabs}
\usepackage{longtable}
\usepackage{amssymb}
\usepackage{float}
\usepackage{wrapfig}
\usepackage{fancyvrb}
\usepackage{xcolor}
\usepackage{hyperref}
\usepackage{url}

\DefineVerbatimEnvironment{promptbox}{Verbatim}{
  frame=single,
  framerule=0.4pt,
  framesep=2mm,
  fontsize=\small,
  xleftmargin=0.02\linewidth,
  xrightmargin=0.02\linewidth
}

\newsavebox{\appendixtablebox}
\newcommand{\appendixtablestyle}{%
  \small
  \setlength{\tabcolsep}{5pt}%
  \renewcommand{\arraystretch}{1.0}%
}
\newcommand{\appendixlongtablestyle}{%
  \small
  \setlength{\LTcapwidth}{\linewidth}%
  \setlength{\tabcolsep}{2.5pt}%
  \renewcommand{\arraystretch}{1.0}%
}

\newenvironment{autowidthtabular}[1]{%
  \begin{lrbox}{\appendixtablebox}%
  \begin{tabular}{#1}%
}{%
  \end{tabular}%
  \end{lrbox}%
  \ifdim\wd\appendixtablebox>\linewidth
    \resizebox{\linewidth}{!}{\usebox{\appendixtablebox}}%
  \else
    \usebox{\appendixtablebox}%
  \fi
}

\title{Learning to Reason with Compressed Context: Ground-Truth-Free Adaptation of OmniLLMs via Self-Distillation}

\author{%
\bfseries Jianghao Wang$^{1,*}$ \quad Ke Meng$^{2}$ \quad
Jian Li$^{2}$ \quad Chi Cheng$^{2}$ \quad Longyu Qi$^{2}$ \\
\bfseries Liyin Liang$^{2}$ \quad Yifeng Qian$^{2}$ \quad
Chunbo Lai$^{2}$ \quad Yutian Lin$^{1,\dagger}$ \quad Zeyu Wang$^{1,\dagger}$ \\
{\normalfont $^{1}$\,School of Computer Science, Wuhan University} \\
{\normalfont $^{2}$\,Didi Chuxing} \\
{\normalfont\texttt{\{jhwang\_cs, yutian.lin, zywang.ai\}@whu.edu.cn}}
}

\iclrfinalcopy
\begin{document}

\maketitle
\begingroup
\renewcommand{\thefootnote}{\fnsymbol{footnote}}
\footnotetext[1]{This work was done during an internship at Didi Chuxing.}
\footnotetext[2]{Corresponding authors.}
\footnotetext[0]{Project page: {\definecolor{projectlink}{RGB}{30,80,150}\hypersetup{pdfborder={0 0 0}}\href{https://github.com/Bamboos2003/CAFD}{\textcolor{projectlink}{\nolinkurl{https://github.com/Bamboos2003/CAFD}}}}.}
\endgroup
\lhead{Preprint.}

\begin{abstract}

Omni-modal large language models (OmniLLMs) enable unified audio-video understanding, but their long multimodal token sequences make deployment computationally expensive. Token compression reduces this cost, yet aggressive compression often lowers accuracy. Existing works predominantly focus on designing better compression mechanisms; however, adapting the underlying language model to reason effectively over the remaining compressed context remains under-explored. To address this, we propose \textbf{\emph{CAFD}} (Compressed-Context Adaptation via Full-Context Distillation), a ground-truth-free self-distillation framework that adapts OmniLLMs to fixed compression pipelines without requiring reference answers, rationales, or correctness rewards. CAFD leverages the full-token view of the same multimodal sample as a source of privileged information: a full-context self-teacher provides soft target supervision to a compressed-context student along the student's on-policy trajectory. Evaluated on Qwen2.5-Omni-7B across five audio-video benchmarks, five compression pipelines, and five deployment budgets, CAFD demonstrates consistent gains, improving 120 out of 125 conditions with an average accuracy boost of 1.44 points and recovering 26.9\% of the accuracy gap on average. These results demonstrate that the proposed ground-truth-free adaptation offers an effective and practical route to improving the accuracy–efficiency trade-off in deployed OmniLLMs.
\end{abstract}

\suppressfloats[t]
\begin{figure}[t]
    \centering
    \includegraphics[width=\linewidth]{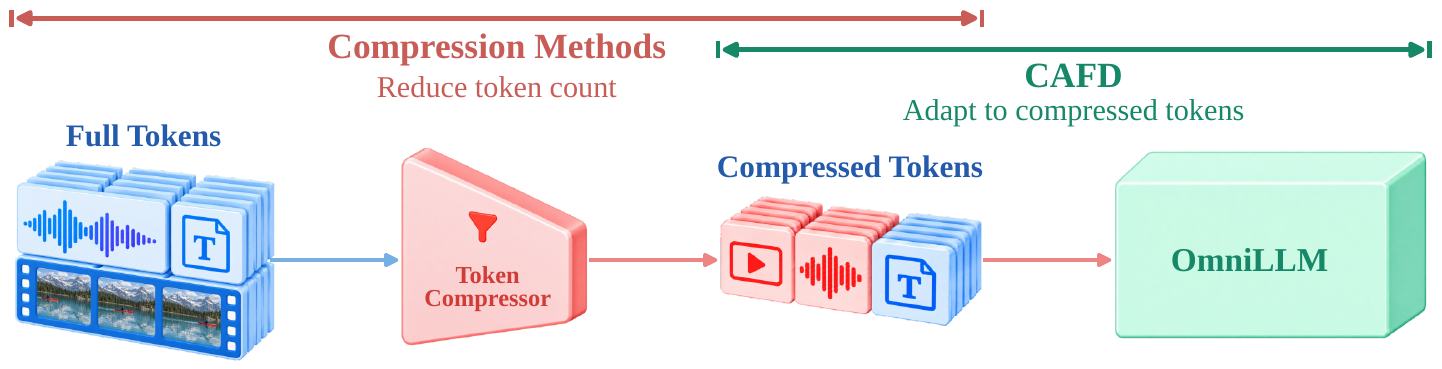}\par
    \makebox[\linewidth][c]{\textbf{(a)}}\par\medskip
    \includegraphics[width=\linewidth]{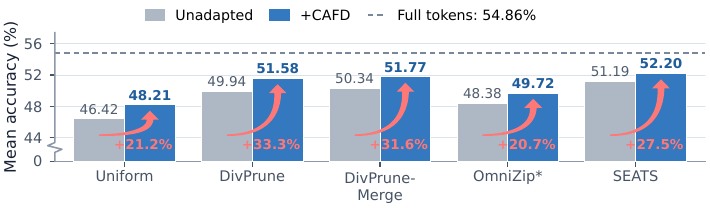}\par
    \makebox[\linewidth][c]{\textbf{(b)}}
    \caption{\textbf{Complementary adaptation and accuracy gains.}
    (a) Compression methods reduce the token count; CAFD complements
    them by adapting the OmniLLM to compressed tokens while keeping the
    compressor unchanged. Pre-LLM compression
    is illustrated; CAFD also supports inner-LLM and hybrid compression.
    (b) Gray and blue bars show each compressor's mean accuracy before and
    after CAFD adaptation, averaging five benchmarks and five
    deployment budgets. The dashed line
    denotes the unadapted full tokens baseline (54.86\%). Red labels indicate accuracy-gap recovery: the percentage
    of the full-token--unadapted accuracy gap closed by adaptation.}
    \label{fig:supervision_comparison}
    \label{fig:aggregate_results}
\end{figure}

\section{Introduction}
\label{sec:intro}

Omni-modal large language models (OmniLLMs) have recently emerged as a unified
interface for reasoning over audio, video, and text~\citep{qwen2_5_omni,qwen3_omni,
vita,video_salmonn,humanomni_v2,omnivinci}. However, long-form audio and video can
produce thousands of multimodal tokens, making prefill computation and memory
increasingly costly. Token compression has therefore become an important
approach to efficient OmniLLM deployment: existing methods prune, merge, select,
or progressively reduce multimodal tokens before or within the language
model~\citep{omnizip,seats,omnifit,omnipack}.

Compression, however, introduces a fundamental trade-off. Reducing the token
budget can remove or weaken task-relevant evidence, leading to an accuracy gap
relative to full-token inference. Existing work largely addresses this problem
by improving the compression mechanism itself, for example, designing better token selection, merging, allocation strategies. This raises
a complementary question that is less explored: \emph{once a compression
pipeline is fixed, can the OmniLLM itself be adapted to better operate on the
information that remains?} A deployed
system may already have a chosen compressor, and improving
the model's ability to use compressed context does not necessarily require
redesigning the compression algorithm or changing its inference architecture.

A natural approach is to adapt the model using additional training data, yet conventional supervision uses reference answers, rationales, evidence annotations, or correctness-derived rewards that require additional labeled data~\citep{visionzip,epic,tbd,o_marc}, which limits the generality of compression adaptation. We instead ask whether the model can be adapted to compressed context without any task-specific labels or reference responses. 
Our key observation is that token compression does not simply produce a different input: it produces \emph{a different information view of the same underlying audio-video sample}. During adaptation, the same
sample can be processed through compressed and full-token paths. The full-token
path has access to information that the deployment model does not, while the
compressed path follows the deployment compression pipeline. This naturally
creates a privileged-information setting: the full-context model provides
soft supervision for learning under compressed context without a reference response.

Based on this observation, we propose \textbf{\emph{CAFD}}
(Compressed-Context Adaptation via Full-Context Distillation), a
ground-truth-free adaptation framework for OmniLLMs under token compression
(Figure~\ref{fig:supervision_comparison}(a)). CAFD constructs a
compressed-context student and a full-context self-teacher from the same
OmniLLM. The student generates a response using the deployment compression
pipeline. The teacher processes the same sample with full tokens and provides
full-vocabulary soft targets along the student's trajectory,
helping the student exploit the retained information. Following OPSD~\citep{opsd},
the supervision is applied to the student's actual response prefixes rather
than to prescribed reference responses. Crucially, the teacher requires no
reference answer, rationale, evidence annotation, or correctness-derived
reward: its additional input information comes from the same sample's full-token
view. The teacher is used only during adaptation and is discarded at deployment, leaving the compressor and inference architecture
unchanged.

We evaluate CAFD on Qwen2.5-Omni-7B across five audio-video benchmarks,
five heterogeneous compression pipelines, and five deployment budgets,
covering token selection, token merging, and progressive inner-LLM pruning.
The results show that adaptation consistently complements compression: with
only 5,700 training instances from 1,485 videos, CAFD improves 120 of
125 matched conditions, increasing average accuracy from 49.26\% to 50.70\%
(+1.44 points). The mean accuracy-gap recovery across the five compressors
is 26.9\% (Figure~\ref{fig:aggregate_results}(b)). At the nominal 5\% retention
budget, the average improvement increases to 1.73 points, with gains in all
25 benchmark--compressor combinations. Notably, each model is adapted once
at 5\% retention and reused across all five deployment budgets.
Controlled experiments show benefits from more training data and strong but
non-extreme training compression, with Jensen--Shannon divergence outperforming
forward and reverse KL. Providing reference answers and rationales to the
teacher does not improve mean accuracy over the proposed ground-truth-free
supervision, demonstrating the effectiveness of supervision from the full-token
view alone.

Our contributions are threefold:
\begin{itemize}
    \item We formulate ground-truth-free adaptation of OmniLLMs to existing
    audio-video token compression pipelines, focusing on learning to use
    compressed context without reference answers or rationales. This complements
    compressor design by addressing how the model uses retained information,
    rather than which tokens the compressor retains.

    \item We introduce an on-policy self-distillation framework in which a
    full-context self-teacher supervises a compressed-context student along the
    student's own response trajectory. The original audio-video sample serves
    as the source of privileged information, eliminating the need to construct
    reference responses or evidence annotations.

    \item Across five compression pipelines, five audio-video benchmarks, and
    five deployment budgets, CAFD improves 120 of 125 matched conditions
    and raises average accuracy by 1.44 points overall, with 26.9\% mean
    accuracy-gap recovery across compressors. The gain reaches 1.73 points at
    nominal 5\% retention. The adapted model can also be reused across deployment budgets
    without retraining.
\end{itemize}

\section{Related Work}
\label{sec:related_work}

\subsection{OmniLLM Token Compression}

OmniLLM compressors operate before the LLM, within it, or at both stages.
Pre-LLM methods~\citep{omnizip,dash,context_guard,omni_refine,omni_select,
omni_focus,remo,omni_prune,omni_scope,omni_delta,echoing_pixels,omni_sift,avoc}
reduce encoder outputs using token saliency, spatiotemporal redundancy,
cross-modal cues, or query relevance.
OMAC~\citep{o_marc} similarly constructs compact audio and visual memory tokens.
Inner-LLM methods~\citep{omnifit,macer} compress intermediate decoder states,
and hybrid pipelines~\citep{omni_drop,seats,omnipack,a_pack} operate at both
stages.

Compressors also differ in their training requirements. Many are
training-free and update no parameters. Some operate directly, whereas OmniFit,
ReMo, OmniDelta, and MACER~\citep{omnifit,remo,omni_delta,macer} require offline
profiling, statistics, skill construction, or calibration.
ContextGuard~\citep{context_guard} trains an auxiliary predictor without
updating the LLM, whereas EchoingPixels, OmniSIFT, and
AVOC~\citep{echoing_pixels,omni_sift,avoc} jointly train compression modules and
the decoder.
Across these groups, the compression mechanism itself remains a primary design
target. Our work complements compressor design by adapting the OmniLLM under
a fixed compression pipeline.

\subsection{On-Policy Self-Distillation}

OPSD~\citep{opsd} samples student trajectories and provides next-token
supervision using a teacher conditioned on privileged solutions.
Vision-OPD~\citep{vision_opd} uses automatically localized evidence crops,
whereas Imagine-OPD~\citep{imagine_opd} uses annotated evidence crops and
ground-truth answers. For multimodal affective computing,
OmniOPSD~\citep{omniopsd_affective} supplies externally generated rationales.
Clue-OPSD~\citep{clue_opsd} uses JSD to supervise full-video student
rollouts with an EMA teacher viewing annotated temporal clue intervals,
without answer labels.
These methods use reference solutions, generated rationales, or localized
evidence to provide teacher privilege. Our work instead uses the original
audio-video input in full, without constructing reference responses or
localizing evidence regions.

RP-OPSD and NOPD~\citep{rp_opsd,nopd} instead pair degraded student images
with higher-quality teacher views without reference answers. Concurrent
S$^2$VOPD~\citep{s2vopd} studies visual augmentations and visual-token dropping
with a clean-view EMA teacher.

\subsection{Adaptation to Compressed Contexts}

VisionZip~\citep{visionzip} and LLaVA-PruMerge~\citep{llava_prumerge} evaluate
supervised instruction tuning after training-free visual-token reduction.
Within audio-video OmniLLMs, O-MARC~\citep{o_marc} builds on GRPO~\citep{grpo}
to adapt to OMAC-compressed inputs using paired full/compressed
answer-and-format rewards.

VoCo-LLaMA~\citep{voco_llama} learns compact representations through
full-to-compressed matching.
EPIC~\citep{epic} supports existing pruning strategies and distills a
more-compressed student from a weight-shared, less-compressed teacher
on prescribed, teacher-forced responses.
TBD~\citep{tbd} adapts compressed-video students across compressors using
LoRA, full-token teacher distillation, and ground-truth answer supervision.
SPIRAL~\citep{spiral} aligns rendered-image Vision-Text Compression with
native-text behavior through on-policy distillation, using training samples
filtered by teacher correctness.

Ground-truth-free on-policy adaptation remains underexplored across
audio-video compression pipelines. CAFD addresses this gap and
demonstrates accuracy gains across token deletion, merging, and
inner-LLM pruning.

\section{Method}
\label{sec:method}

\begin{figure}[t]
    \centering
    \includegraphics[width=\linewidth]{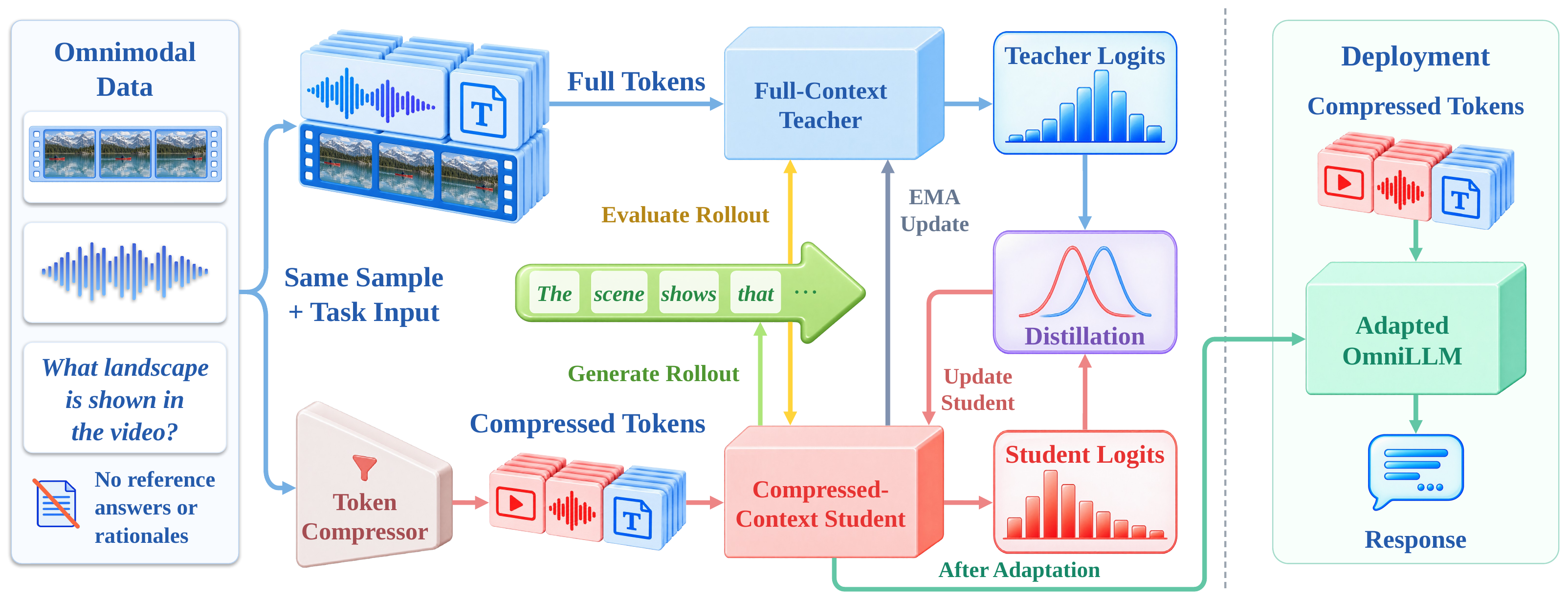}
    \caption{\textbf{Overview of CAFD.} Self-distillation guides the model
    under compressed context toward its full-context capabilities.
    The compressed student generates a response; both paths compute next-token
    distributions along it using the same sample and instruction, without
    reference answers or rationales. Gradients update only the student;
    the teacher parameters track the student via an exponential moving average (EMA).
    Deployment retains the adapted model and original compressor.
    The framework supports pre-LLM (illustrated), inner-LLM, and hybrid compression.}
    \label{fig:cafd_overview}
\end{figure}

In this section, we present CAFD, which adapts an OmniLLM to compressed
audio-video inputs through self-distillation without reference answers or
rationales (Figure~\ref{fig:cafd_overview}). We formulate the problem, construct
a full-context self-teacher, and describe the on-policy distillation objective
and parameter updates.

\subsection{Problem Formulation}

OmniLLM token compression reduces computational cost by selecting, merging,
or pruning audio-video tokens before or within the LLM.
CAFD complements these methods by adapting the model to their compressed
contexts while keeping the compression pipeline unchanged.

Let \(\mathcal{D}=\{(x_i,u_i)\}_{i=1}^{N}\) denote a training
corpus, where \(x_i=(x_i^{a},x_i^{v})\) is an audio-video sample and
\(u_i\) is a textual instruction. We seek to adapt a pretrained OmniLLM by
optimizing its trainable model parameters \(\theta\) under the chosen
fine-tuning parameterization.

Let \(\pi^{\mathrm{comp}}_{m,r}\) denote the compressed-context path using compressor \(m\)
at budget configuration \(r\), and let \(\pi^{\mathrm{full}}\) denote the
full-context path. This notation covers pre-LLM, inner-LLM, and hybrid compression.
For a given \(m,r\), adaptation updates model parameters while token
selection and allocation follow the compressor's own rules.
For either path \(\pi\), we write the next-token distribution given response
prefix \(y_{<t}\) as \(p_{\theta}(\cdot\mid x,u,\pi,y_{<t})\), and the
complete-response distribution as \(p_{\theta}(y\mid x,u,\pi)\).
Our objective is to improve the model under \(\pi^{\mathrm{comp}}_{m,r}\)
without redesigning the compressor.

\subsection{Full-Context Self-Teacher}
\label{sec:full_to_compressed}

The student and self-teacher start from the same pretrained OmniLLM.
For each \((x,u)\in\mathcal{D}\), the student with parameters \(\theta\) follows
\(\pi^{\mathrm{comp}}_{m,r_{\mathrm{train}}}\), where \(r_{\mathrm{train}}\)
is the training budget configuration. The teacher processes the same sample
through \(\pi^{\mathrm{full}}\) using parameters \(\bar{\theta}\), an exponential
moving average of \(\theta\), initialized as \(\bar{\theta}_0=\theta_0\).
At a shared prefix \(y_{<t}\), their next-token distributions are
\begin{equation}
    \begin{aligned}
        p_t(\cdot)
        &=p_{\theta}
          \bigl(\cdot\mid x,u,\pi^{\mathrm{comp}}_{m,r_{\mathrm{train}}},y_{<t}\bigr),\\
        q_t(\cdot)
        &=p_{\bar{\theta}}
          \bigl(\cdot\mid x,u,\pi^{\mathrm{full}},y_{<t}\bigr).
    \end{aligned}
    \label{eq:paired_distributions}
\end{equation}
Both paths use the same instruction. The teacher provides full-vocabulary
supervision with access to the sample's full audio-video context, whereas the
student uses compressed context. The teacher receives no reference answer or rationale, evidence
annotation, or correctness-derived reward.

\subsection{On-Policy Self-Distillation Objective}
\label{sec:trajectory_distillation}

Compression can change how the student generates its response. We
therefore sample an on-policy response trajectory from the current
compressed student:
\begin{equation}
    \hat{y}^{S}
    \sim \tilde{p}_{\theta}
    \bigl(\cdot\mid x,u,\pi^{\mathrm{comp}}_{m,r_{\mathrm{train}}}\bigr).
    \label{eq:on_policy_rollout}
\end{equation}
Here \(\tilde{p}\) applies the decoding settings in
Appendix~\ref{app:training_configuration} to the student's distribution.
Conditioned on each student-generated prefix \(\hat{y}^{S}_{<t}\), the student
and teacher compute the full-vocabulary next-token distributions \(p_t\) and
\(q_t\) in Equation~\ref{eq:paired_distributions}.

We use Jensen--Shannon divergence~\citep{jsd} to match the two distributions.
For each vocabulary token \(v\in\mathcal{V}\), define the mixture and
divergence contribution as
\begin{equation}
    \begin{aligned}
        m_t(v)&=\tfrac12\bigl[p_t(v)+q_t(v)\bigr],\\
        d_{t,v}&=\tfrac12 p_t(v)\log\frac{p_t(v)}{m_t(v)}
                  +\tfrac12 q_t(v)\log\frac{q_t(v)}{m_t(v)}.
    \end{aligned}
    \label{eq:jsd_contribution}
\end{equation}
Following OPSD~\citep{opsd}, we retain its pointwise clipping rule, which caps
each vocabulary contribution at \(\delta\). For a completion of length \(T_y\),
the objective is
\begin{equation}
    \mathcal{L}_{\mathrm{adapt}}
    =\frac{1}{T_y}\sum_{t=1}^{T_y}\sum_{v\in\mathcal{V}}
      \min\!\left(d_{t,v},\delta\right),
    \qquad \delta=0.05.
    \label{eq:clipped_jsd}
\end{equation}
Clipping is applied to the individual loss terms before summing over the
vocabulary; it does not truncate or renormalize either probability distribution.
Contributions above \(\delta\) are capped at \(\delta\) and have zero local
gradient through the clipping operation.
We average the resulting loss over generated response positions and exclude
prompt positions. Gradients flow through the student's next-token distributions only; the sampled
trajectory and teacher probabilities remain detached.

After each successful student optimizer step \(k\), we update
\(\bar{\theta}_k=\mu\bar{\theta}_{k-1}+(1-\mu)\theta_k\), with
\(\mu=0.999\). The teacher forward pass for that step uses
\(\bar{\theta}_{k-1}\).

We adapt the model separately for each compressor at \(r_{\mathrm{train}}\).
Deployment discards the teacher and uses the updated model parameters with
the original inference architecture and compressor at the chosen deployment budget.
Whether token counts and computation change depends on the compressor:
fixed-count selection preserves token counts, whereas model-dependent pruning
can alter intermediate-layer token counts after adaptation
(Appendix~\ref{app:compression_results}).

\section{Experiments}
\label{sec:experiments}

In this section, we conduct extensive experiments across five audio-video
benchmarks, five compression pipelines, and five deployment budgets to
demonstrate the effectiveness of CAFD. Beyond this broad evaluation,
controlled studies examine training data scale, training retention,
distillation objectives, and supervision sources. Together, these experiments
demonstrate broad accuracy improvements and establish CAFD as a practical
approach to adapting compressed OmniLLMs without reference answers or rationales.

\subsection{Experimental Settings}
\label{sec:experimental_setup}

\paragraph{Backbone and benchmarks.}
We use Qwen2.5-Omni-7B~\citep{qwen2_5_omni} and evaluate multiple-choice
question-answering accuracy on WorldSense~\citep{worldsense} (3,172
questions), DailyOmni~\citep{dailyomni} (1,197), AVUT~\citep{avut} (1,734),
OmniVideoBench~\citep{omnivideobench} (1,000), and
Video-MME~\citep{video_mme} (2,700).
Appendices~\ref{app:benchmarks} and~\ref{app:media_preparation} describe the datasets and their local media
preparation.

\paragraph{Target compression pipelines.}
We evaluate five pipelines: Uniform, DivPrune~\citep{divprune},
DivPrune-Merge, OmniZip*~\citep{omnizip}, and SEATS~\citep{seats}.
The first four compress tokens before the LLM, with DivPrune-Merge and
OmniZip* incorporating token merging. SEATS combines pre-LLM selection with
inner-LLM pruning. Appendices~\ref{app:compression_details}
and~\ref{app:budget_configurations} detail the implementations, including
our OmniZip* variant, and budget configurations.

\paragraph{Training configuration.}
We use 5,700 instances from 1,485 OmniVideo~\citep{omnivideo_100k} videos,
balanced across ten task types. Unless stated otherwise, training uses
the media, question, and all candidate options,
without reference answers or rationales.

Training videos are sampled at a target rate of 2 FPS and capped at 128
frames. For training efficiency, we use rank-64 LoRA~\citep{lora} on the Thinker decoder's
attention and MLP projections using AdamW~\citep{adamw}, a learning rate of
\(5\times10^{-6}\), and a global batch size of 32 on 8 NVIDIA RTX 6000D GPUs.
Appendices~\ref{app:prompts} and~\ref{app:training_configuration} provide
the training prompt templates and the full set of optimization and rollout
settings, respectively.

\paragraph{Evaluation protocol.}
For the shorter-video benchmarks (WorldSense, DailyOmni, and AVUT), we use
up to 128 frames and full audio tracks to match the video's temporal coverage.
For the longer-video benchmarks (OmniVideoBench and Video-MME), we use up to
256 and 768 frames, respectively, and retain the first 300 seconds of audio
following SEATS~\citep{seats}. Frames are sampled across the full video at a
target rate of 2 FPS, using the same per-frame pixel bounds as training
(Appendix~\ref{app:training_configuration}). Subtitles are excluded.
Appendix~\ref{app:evaluation_protocol} provides further evaluation details.

\paragraph{Comparison setup.}
For each compressor, we compare the original model (\emph{Unadapted}) with
its adapted version (\emph{+CAFD}) under the same compression
configuration, using the original full-token model as a reference. Each model is trained at 5\%
retention and evaluated at 35\%, 25\%, 15\%, 10\%, and 5\% with the same
compressor, without retraining. For each compressor and nominal budget,
compression hyperparameters are fixed across all five benchmarks and
before and after adaptation; each matched pair differs only in Thinker weights.

Retention budgets are nominal; realized retention and computation are reported
in Table~\ref{tab:retention_flops}. Appendices~\ref{app:budget_configurations}
and~\ref{app:comparison_setup} detail compression settings and comparison
protocols, respectively. Statistics are rounded only for reporting, so
calculations from displayed values may differ slightly.

\subsection{Main Results}
\label{sec:main_results}

\begin{table}[t]
    \caption{\textbf{Main results across compressors, budgets, and benchmarks.}
    Each cell reports \emph{Unadapted / +CAFD} accuracy (\%). Each
    +CAFD model is trained under the
    corresponding compressor's 5\% budget configuration and is evaluated at the
    five listed budgets. Column Avg.
    averages benchmarks, block-ending Avg. averages budgets, and their
    intersection averages all 25 conditions. The higher value in each pair is
    bold. DivPrune-M denotes DivPrune-Merge.}
    \label{tab:main_results}
    \centering
    \scriptsize
    \setlength{\tabcolsep}{3.1pt}
    \resizebox{\linewidth}{!}{%
    \begin{tabular}{llcccccc}
        \toprule
        Compressor & Budget & WorldSense & DailyOmni & AVUT &
        OmniVideoBench & Video-MME & Avg. \\
        \midrule
        Full tokens & 100\% & 46.85 & 62.91 & 64.65 & 35.50 & 64.41 & 54.86 \\
\midrule
Uniform & 35\% & 43.28 / \textbf{44.96} & 57.23 / \textbf{58.65} & 60.44 / \textbf{62.86} & \textbf{33.80} / 33.70 & 63.44 / \textbf{64.81} & 51.64 / \textbf{53.00} \\
 & 25\% & 41.96 / \textbf{43.66} & 52.05 / \textbf{55.89} & 56.81 / \textbf{59.05} & 31.40 / \textbf{32.60} & 61.63 / \textbf{62.70} & 48.77 / \textbf{50.78} \\
 & 15\% & 38.84 / \textbf{39.97} & 49.37 / \textbf{52.72} & 52.94 / \textbf{55.25} & 29.90 / \textbf{31.50} & 60.33 / \textbf{61.67} & 46.28 / \textbf{48.22} \\
 & 10\% & 36.44 / \textbf{37.64} & 45.45 / \textbf{48.45} & 50.35 / \textbf{51.50} & 29.50 / \textbf{30.80} & 58.15 / \textbf{60.22} & 43.98 / \textbf{45.72} \\
 & 5\% & 35.18 / \textbf{36.66} & 40.77 / \textbf{43.69} & 46.37 / \textbf{47.92} & 30.10 / \textbf{31.00} & 54.89 / \textbf{57.37} & 41.46 / \textbf{43.33} \\
\cmidrule(l){2-8}
 & Avg. & 39.14 / \textbf{40.58} & 48.97 / \textbf{51.88} & 53.38 / \textbf{55.32} & 30.94 / \textbf{31.92} & 59.69 / \textbf{61.36} & 46.42 / \textbf{48.21} \\
\midrule
DivPrune & 35\% & 45.49 / \textbf{46.37} & 58.48 / \textbf{60.82} & 62.28 / \textbf{63.21} & 34.90 / \textbf{35.50} & 63.15 / \textbf{64.44} & 52.86 / \textbf{54.07} \\
 & 25\% & 45.02 / \textbf{46.15} & 56.73 / \textbf{58.73} & 60.90 / \textbf{62.00} & 34.30 / \textbf{35.00} & 62.04 / \textbf{63.15} & 51.80 / \textbf{53.01} \\
 & 15\% & 43.25 / \textbf{44.45} & 55.30 / \textbf{58.06} & 58.19 / \textbf{60.03} & 32.30 / \textbf{34.70} & 61.15 / \textbf{62.89} & 50.04 / \textbf{52.03} \\
 & 10\% & 42.06 / \textbf{43.35} & 53.55 / \textbf{55.22} & 56.69 / \textbf{58.30} & 33.60 / \textbf{35.30} & 59.56 / \textbf{61.04} & 49.09 / \textbf{50.64} \\
 & 5\% & 39.12 / \textbf{40.45} & 49.62 / \textbf{52.38} & 52.88 / \textbf{54.27} & 31.50 / \textbf{34.20} & 56.48 / \textbf{59.52} & 45.92 / \textbf{48.16} \\
\cmidrule(l){2-8}
 & Avg. & 42.99 / \textbf{44.16} & 54.74 / \textbf{57.04} & 58.19 / \textbf{59.56} & 33.32 / \textbf{34.94} & 60.47 / \textbf{62.21} & 49.94 / \textbf{51.58} \\
\midrule
DivPrune-M & 35\% & 45.40 / \textbf{45.68} & 59.06 / \textbf{61.15} & 62.46 / \textbf{63.44} & 34.70 / \textbf{35.70} & 63.30 / \textbf{64.33} & 52.98 / \textbf{54.06} \\
 & 25\% & 44.45 / \textbf{45.81} & 57.73 / \textbf{58.90} & 61.36 / \textbf{62.98} & 34.20 / \textbf{35.40} & 62.41 / \textbf{64.00} & 52.03 / \textbf{53.42} \\
 & 15\% & 43.69 / \textbf{44.99} & 55.39 / \textbf{58.23} & 59.52 / \textbf{60.09} & 33.50 / \textbf{34.70} & 61.11 / \textbf{62.63} & 50.64 / \textbf{52.13} \\
 & 10\% & 42.72 / \textbf{44.04} & 53.30 / \textbf{56.31} & 57.09 / \textbf{59.11} & 33.90 / \textbf{34.60} & 59.96 / \textbf{61.74} & 49.39 / \textbf{51.16} \\
 & 5\% & 39.47 / \textbf{41.05} & 50.29 / \textbf{51.96} & 53.92 / \textbf{55.25} & 32.50 / \textbf{32.70} & 57.07 / \textbf{59.48} & 46.65 / \textbf{48.09} \\
\cmidrule(l){2-8}
 & Avg. & 43.15 / \textbf{44.31} & 55.15 / \textbf{57.31} & 58.87 / \textbf{60.17} & 33.76 / \textbf{34.62} & 60.77 / \textbf{62.44} & 50.34 / \textbf{51.77} \\
\midrule
OmniZip* & 35\% & 44.04 / \textbf{45.71} & 59.15 / \textbf{60.65} & 61.13 / \textbf{62.40} & \textbf{35.00} / 34.70 & 63.07 / \textbf{64.26} & 52.48 / \textbf{53.54} \\
 & 25\% & 43.25 / \textbf{44.58} & 57.64 / \textbf{58.81} & 59.23 / \textbf{60.09} & 31.80 / \textbf{34.00} & 62.89 / \textbf{63.74} & 50.96 / \textbf{52.24} \\
 & 15\% & 39.97 / \textbf{41.24} & 54.80 / \textbf{57.81} & 55.59 / \textbf{56.75} & 33.00 / \textbf{33.30} & 61.04 / \textbf{61.81} & 48.88 / \textbf{50.18} \\
 & 10\% & 38.75 / \textbf{40.64} & 50.38 / \textbf{52.97} & 54.44 / \textbf{54.90} & 31.70 / \textbf{32.80} & 59.04 / \textbf{60.56} & 46.86 / \textbf{48.37} \\
 & 5\% & 34.93 / \textbf{36.48} & 45.36 / \textbf{47.62} & 47.75 / \textbf{48.67} & 30.50 / \textbf{31.60} & 55.04 / \textbf{56.93} & 42.72 / \textbf{44.26} \\
\cmidrule(l){2-8}
 & Avg. & 40.19 / \textbf{41.73} & 53.47 / \textbf{55.57} & 55.63 / \textbf{56.56} & 32.40 / \textbf{33.28} & 60.21 / \textbf{61.46} & 48.38 / \textbf{49.72} \\
\midrule
SEATS & 35\% & 46.44 / \textbf{47.51} & 61.32 / \textbf{62.24} & 64.94 / \textbf{65.97} & \textbf{35.70} / 35.50 & 65.44 / \textbf{66.19} & 54.77 / \textbf{55.48} \\
 & 25\% & 45.84 / \textbf{47.32} & 59.90 / \textbf{60.48} & 62.69 / \textbf{63.55} & \textbf{35.40} / 35.00 & 64.85 / \textbf{65.96} & 53.74 / \textbf{54.46} \\
 & 15\% & 44.58 / \textbf{45.62} & 56.64 / \textbf{58.65} & 59.52 / \textbf{60.84} & \textbf{36.00} / 35.70 & 63.26 / \textbf{64.22} & 52.00 / \textbf{53.01} \\
 & 10\% & 43.32 / \textbf{44.10} & 56.31 / \textbf{56.89} & 57.90 / \textbf{58.77} & 34.30 / \textbf{36.20} & 61.48 / \textbf{62.48} & 50.66 / \textbf{51.69} \\
 & 5\% & 37.89 / \textbf{39.31} & 48.71 / \textbf{50.63} & 49.65 / \textbf{51.61} & 31.20 / \textbf{31.90} & 56.44 / \textbf{58.30} & 44.78 / \textbf{46.35} \\
\cmidrule(l){2-8}
 & Avg. & 43.61 / \textbf{44.77} & 56.57 / \textbf{57.78} & 58.94 / \textbf{60.15} & 34.52 / \textbf{34.86} & 62.30 / \textbf{63.43} & 51.19 / \textbf{52.20} \\
\bottomrule
    \end{tabular}}
\end{table}

Table~\ref{tab:main_results} compares accuracy before and after CAFD
adaptation across five compressors, five retention budgets, and five benchmarks.
The evaluation spans token
deletion, token merging, and progressive inner-LLM compression, while holding
the compressor and deployment budget configuration fixed within every pair.
CAFD improves 120 of the 125 matched cells. Across all 125 conditions,
average accuracy rises from 49.26\% to 50.70\%, a gain of
1.44 percentage points. At the nominal 5\% evaluation budget, average accuracy across
the five compressors and five benchmarks increases from 44.31\% to 46.04\%,
a gain of 1.73 percentage points, with gains in all 25 conditions.
For each compressor, Figure~\ref{fig:aggregate_results}(b) divides its mean
adaptation gain by its full-token--Unadapted accuracy gap. The five ratios
average to 26.9\% accuracy-gap recovery.

The average gains at the 35\%, 25\%, 15\%, 10\%, and 5\% budget labels are
1.08, 1.32, 1.54, 1.52, and 1.73 points, respectively.
These results demonstrate that CAFD effectively improves compressed
inference across diverse compression pipelines, with the largest average gain
at 5\% retention. These gains are achieved by training one model per
compressor and reusing it across all five deployment budgets without retraining.

Adaptation also improves the accuracy--budget trade-off. Averaged across the
five benchmarks, adapted DivPrune at 25\% retention reaches 53.01\% accuracy,
exceeding the unadapted model's 52.86\% at 35\%. Similarly, adapted
DivPrune-Merge at 15\% reaches 52.13\%, compared with 52.03\% for its
unadapted counterpart at 25\%. These comparisons show that CAFD can
preserve or improve average accuracy at a lower nominal retention budget,
making more aggressive compression a practical deployment option.

All five compressors benefit in aggregate, from Uniform sampling and
diversity-based selection to token merging and progressive inner-LLM pruning.
In particular, improvements with DivPrune-Merge, OmniZip*, and SEATS show
that adaptation remains useful alongside compression strategies that already
aggregate token information or reduce tokens progressively. This breadth
supports CAFD as a complementary approach to improving existing
compression pipelines through model adaptation.
Appendix~\ref{app:compression_results} provides detailed retention and compute
results; Appendix~\ref{app:behavior_details} analyzes prediction changes,
correct-option scores, and task-level results.

\subsection{Additional Experiments}
\label{sec:design_controls}

\paragraph{Training data scale.}
\label{sec:training_conditions}

\begin{figure}[htbp]
    \centering
    \includegraphics[width=\linewidth]{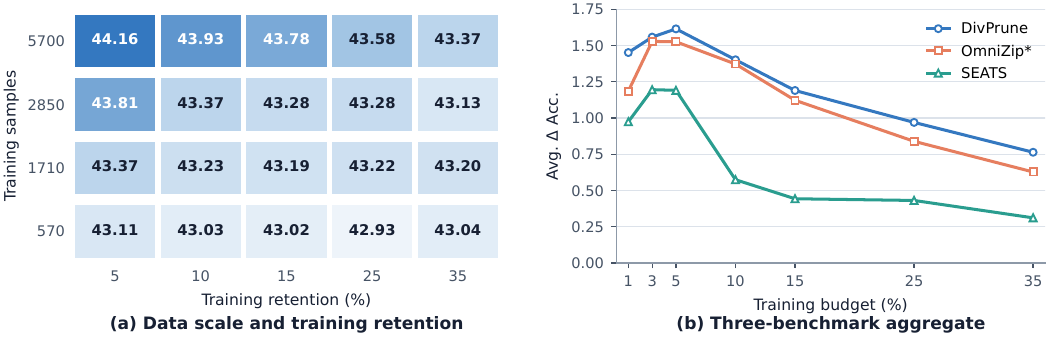}
    \caption{\textbf{Effects of training data scale and retention.}
    (a) WorldSense five-budget average accuracy for DivPrune across training-set
    sizes and training retentions. (b) Equal-weight accuracy-point gain over
    the matched Unadapted baseline across WorldSense, DailyOmni, and AVUT at
    five deployment budgets per benchmark.
    All runs use two epochs; data scale therefore changes both the number of
    training examples and the number of optimizer updates.}
    \label{fig:train_rtt_and_datacount}
\end{figure}

Figure~\ref{fig:train_rtt_and_datacount}(a) shows that more data is generally
beneficial for DivPrune across the tested training retentions. At 5\% training
retention, increasing the training set from 570 to 5,700 instances raises
WorldSense accuracy averaged over five deployment budgets from 43.11\% to
44.16\%. At this training retention, every tested data scale improves over
Unadapted, with accuracy increasing at each larger data scale.
Using all 5,700 training instances gives the best accuracy at every tested
training retention, while 5\% training retention performs best at every
data scale. Thus, the benefit of more training data holds across the
tested retention settings, and the strong performance of 5\% training
retention is already apparent with smaller datasets.
Together, these results show that CAFD already benefits
from small training sets and gains further accuracy as training data increases. With epochs fixed, increasing data size also increases optimizer
updates; the trend reflects the combined effect of more data and training.
Appendix~\ref{app:training_condition_details}
(Table~\ref{tab:data_scale_details}) reports accuracy at each deployment budget
for all training-data settings.

\paragraph{Training retention.}
Figure~\ref{fig:train_rtt_and_datacount}(b) compares training retentions by
average gains across three benchmarks and five deployment budgets.
DivPrune has its highest measured mean gain at 5\%, while OmniZip* and SEATS
are nearly tied at 3\% and 5\%. Reducing training retention to 1\% does not
improve these gains further. All three compressors therefore favor training
with relatively strong compression, but the gains do not keep increasing
as the training budget shrinks. One possible explanation is that stronger
compression encourages the student to make better use of limited context,
whereas extremely sparse inputs make the teacher's predictions harder to
learn from the retained evidence. The observed gains near 3--5\% support
5\% as a common training setting for effective adaptation across the tested
deployment budgets.
Appendix~\ref{app:training_condition_details}
(Table~\ref{tab:train_retention_details}) reports the individual benchmark
results at each training and deployment budget.

\paragraph{Distillation objective.}
With the remaining training configuration fixed, JSD outperforms both
forward and reverse KL on all three benchmarks at every evaluated retention
budget (Table~\ref{tab:distillation_design}). This consistent advantage
supports JSD as an effective objective for adapting OmniLLMs to compressed
inputs. With the same training data and compression setting, changing the
distribution-matching objective changes how effectively the student learns
from full-token supervision. The JSD gains extend across deployment budgets,
supporting its use for a model that is trained once and reused at
different retention settings.
Appendix~\ref{app:objective_details} provides the per-budget results.

\begin{table}[!ht]
    \caption{\textbf{Comparison of distillation objectives.}
    All models are trained with DivPrune at 5\% retention and evaluated
    with the same compressor. Accuracies (\%) are averaged over the five
    deployment budgets within each benchmark.}
    \label{tab:distillation_design}
    \centering
    \small
    \setlength{\tabcolsep}{5pt}
    \begin{autowidthtabular}{lcccc}
        \toprule
        Objective & WorldSense & DailyOmni & AVUT & Avg. \\
        \midrule
        \textbf{JSD (CAFD)} & \textbf{44.16} & \textbf{57.04} & \textbf{59.56} & \textbf{53.59} \\
Forward KL & 42.40 & 54.87 & 58.29 & 51.86 \\
Reverse KL & 43.25 & 55.37 & 58.97 & 52.53 \\
\bottomrule
    \end{autowidthtabular}
\end{table}

\paragraph{Task conditioning and supervision.}
\label{sec:task_conditioning}

Table~\ref{tab:supervision_controls} compares CAFD against two controls that make use of reference answers and rationales: one supplies them to the teacher, and the other uses them as training targets for compressed-input SFT. All adaptation runs share the same training data and student prompt. While compressed-input SFT attains the highest average accuracy, it depends on reference responses as supervision targets. CAFD, by contrast, achieves comparable performance without access to any reference answers or rationales, drawing its gains instead from the full-token teacher's distributions. This suggests that reference answers and rationales are not a prerequisite for effective compression adaptation. Moreover, providing such references to the teacher yields no improvement in three-benchmark average accuracy over CAFD.
Additional experiments show that adaptation also remains effective with only media and a shared generic
instruction (Appendix~\ref{app:task_conditioning}).
Appendix~\ref{app:control_details} provides per-budget results for
Table~\ref{tab:supervision_controls}.

\begin{table}[!ht]
    \caption{\textbf{Comparison of supervision sources.}
    All adapted models are trained with DivPrune at 5\% retention; all rows are
    evaluated with DivPrune. Accuracies (\%) are averaged over five
    deployment budgets. ``Reference'' denotes ground-truth answers and rationales.}
    \label{tab:supervision_controls}
    \centering
    \small
    \setlength{\tabcolsep}{5pt}
    \begin{autowidthtabular}{lccccc}
        \toprule
        Method & Reference use & WorldSense & DailyOmni &
        AVUT & Avg. \\
        \midrule
        Unadapted & -- & 42.99 & 54.74 & 58.19 & 51.97 \\
CAFD & None & 44.16 & \underline{57.04} & \underline{59.56} & \underline{53.59} \\
CAFD + reference & Teacher input & \underline{44.17} & 56.66 & 59.49 & 53.44 \\
Compressed-input SFT & Student target & \textbf{44.31} & \textbf{57.78} & \textbf{59.90} & \textbf{54.00} \\
\bottomrule
    \end{autowidthtabular}
\end{table}

\section{Limitations and Future Work}
\label{sec:limitations}

Due to limited computing resources, training currently uses a small dataset and a fixed retention budget for each
model. Future work could investigate whether larger training datasets and
mixed-retention training improve robustness across deployment budgets.

Another direction is to investigate how teacher reliability and the evidence retained after compression can inform the design of distillation objectives. A longer-term direction is to jointly optimize compression and adaptation under a fixed compute budget.

\section{Conclusion}
\label{sec:conclusion}

In this work, we introduced CAFD, a ground-truth-free adaptation framework that enables OmniLLMs to better reason over compressed multimodal context without modifying the compression pipeline or requiring reference responses. By viewing full-token inference as a privileged information source, CAFD uses a full-context self-teacher to distill soft supervision onto a compressed-context student along its own response trajectory, allowing the model to learn how to exploit the information preserved after compression. Extensive experiments across five compression pipelines, five audio-video benchmarks, and five deployment budgets demonstrate that CAFD consistently recovers a substantial portion of the accuracy lost to compression, while a single model adapted at one compression budget can generalize across different deployment budgets. These results highlight that efficient multimodal inference is not only a problem of retaining the right tokens, but also of learning to reason effectively with the information that remains. We hope this perspective motivates further work on adapting OmniLLMs to resource-constrained and information-limited inference settings.

\section*{Acknowledgments}

This work was supported by the CCF-DiDi GAIA Collaborative Research Funds for Young Scholars.

\section*{AI Use Statement}

We used generative AI tools to assist with literature review, research hypothesis refinement, experimental design, implementation, and interpretation of results, as well as manuscript drafting and editing and figure preparation. We reviewed all AI-assisted work and checked reported numerical results against experiment records. We take responsibility for the final content of this paper, including its text, scientific claims, citations, analyses, and figures.

\section*{Reproducibility Statement}

Section~\ref{sec:method} specifies the adaptation framework and training
objective. Appendix~\ref{app:reproducibility} documents the datasets and data
preparation, compressor implementations and budget configurations, training
prompts and hyperparameters, and evaluation and aggregation protocols.
Appendix~\ref{app:detailed_results} provides detailed experimental results
complementing the main-text summaries.

\bibliography{iclr2027_conference}

\begin{thebibliography}{52}
\providecommand{\natexlab}[1]{#1}
\providecommand{\url}[1]{\texttt{#1}}
\expandafter\ifx\csname urlstyle\endcsname\relax
  \providecommand{\doi}[1]{doi: #1}\else
  \providecommand{\doi}{doi: \begingroup \urlstyle{rm}\Url}\fi

\bibitem[Alvar et~al.(2025)Alvar, Singh, Akbari, and Zhang]{divprune}
Saeed~Ranjbar Alvar, Gursimran Singh, Mohammad Akbari, and Yong Zhang.
\newblock Divprune: Diversity-based visual token pruning for large multimodal models.
\newblock In \emph{Proceedings of the IEEE/CVF Conference on Computer Vision and Pattern Recognition (CVPR)}, pp.\  9392--9401, June 2025.

\bibitem[Cai et~al.(2026{\natexlab{a}})Cai, Fu, Zhang, He, and Shan]{omnivideo_100k}
Xinyue Cai, Chaoyou Fu, Yi-Fan Zhang, Ran He, and Caifeng Shan.
\newblock Omnivideo-100k: A dataset for audio-visual reasoning through structured scripts and evidence chains, 2026{\natexlab{a}}.
\newblock URL \url{https://arxiv.org/abs/2606.14702}.

\bibitem[Cai et~al.(2026{\natexlab{b}})Cai, Liu, Liu, Deng, Yao, Zheng, Ouyang, Li, Wang, Sun, Bai, and Li]{imagine_opd}
Yishuo Cai, Jiahui Liu, Yuanxin Liu, Haobo Deng, Linli Yao, Yuhao Zheng, Kun Ouyang, Zhimo Li, Ziyue Wang, Xu~Sun, Haoli Bai, and Xiaohui Li.
\newblock Thinking without images: Internalizing visual manipulation with on-policy self-distillation, 2026{\natexlab{b}}.
\newblock URL \url{https://arxiv.org/abs/2606.08719}.

\bibitem[Cao et~al.(2026)Cao, Zhang, Yu, Zhang, Cao, Lu, Lin, Han, and Sun]{omni_focus}
Shijie Cao, Qingyu Zhang, Boxi Yu, Yuzhong Zhang, Boxi Cao, Yaojie Lu, Hongyu Lin, Xianpei Han, and Le~Sun.
\newblock Omnifocus: Query-guided modality-balanced token compression for omni-modal large language models, 2026.
\newblock URL \url{https://arxiv.org/abs/2607.03050}.

\bibitem[Chen et~al.(2026)Chen, Tan, Yu, Wang, Cheng, Guan, Jiang, Li, Zhu, and Song]{avoc}
Yijing Chen, Wenhui Tan, Xiaoyi Yu, Yuyue Wang, Xin Cheng, Kaisi Guan, Hao Jiang, Xiangyang Li, Guojie Zhu, and Ruihua Song.
\newblock Avoc: Enhancing hour-level audio-video understanding in omni-modal llms via retrieval-inspired token compression, 2026.
\newblock URL \url{https://arxiv.org/abs/2606.24286}.

\bibitem[Cheng et~al.(2026)Cheng, Chen, Yang, Guan, Chen, Lian, Peng, Ma, Cui, and Tian]{omniopsd_affective}
Zebang Cheng, Shuimu Chen, Boxue Yang, Yuanshen Guan, Jingyi Chen, Zheng Lian, Xiaojiang Peng, Fei Ma, LaiZhong Cui, and Qi~Tian.
\newblock Omniopsd: Rationale-privileged on-policy self-distillation for affective computing, 2026.
\newblock URL \url{https://arxiv.org/abs/2606.15920}.

\bibitem[Dao(2024)]{flashattn2}
Tri Dao.
\newblock Flashattention-2: Faster attention with better parallelism and work partitioning.
\newblock In B.~Kim, Y.~Yue, S.~Chaudhuri, K.~Fragkiadaki, M.~Khan, and Y.~Sun (eds.), \emph{International Conference on Learning Representations}, volume 2024, pp.\  35549--35562, 2024.
\newblock URL \url{https://proceedings.iclr.cc/paper_files/paper/2024/file/98ed250b203d1ac6b24bbcf263e3d4a7-Paper-Conference.pdf}.

\bibitem[Deng et~al.(2026)Deng, Cai, Zheng, Wang, Yang, and Han]{omni_refine}
Yuchen Deng, Zidang Cai, Hai-Tao Zheng, Jie Wang, Feidiao Yang, and Yuxing Han.
\newblock Omnirefine: Alignment-aware cooperative compression for efficient omnimodal large language models, 2026.
\newblock URL \url{https://arxiv.org/abs/2605.12056}.

\bibitem[Ding et~al.(2026)Ding, Ji, Li, Liu, Chen, Wu, Li, Zeng, Shi, Guan, Zhang, Liu, Liu, Wan, and Wang]{omni_sift}
Yue Ding, Yiyan Ji, Jungang Li, Xuyang Liu, Xinlong Chen, Junfei Wu, Bozhou Li, Bohan Zeng, Yang Shi, Yushuo Guan, Yuanxing Zhang, Jiaheng Liu, Qiang Liu, Pengfei Wan, and Liang Wang.
\newblock Omni{SIFT}: Modality-asymmetric token compression for efficient omni-modal large language models.
\newblock In \emph{Forty-third International Conference on Machine Learning}, 2026.
\newblock URL \url{https://openreview.net/forum?id=aPnQpmwHW7}.

\bibitem[Fu et~al.(2025{\natexlab{a}})Fu, Dai, Luo, Li, Ren, Zhang, Wang, Zhou, Shen, Zhang, Chen, Li, Lin, Zhao, Li, Xu, Zheng, Chen, Shan, He, and Sun]{video_mme}
Chaoyou Fu, Yuhan Dai, Yongdong Luo, Lei Li, Shuhuai Ren, Renrui Zhang, Zihan Wang, Chenyu Zhou, Yunhang Shen, Mengdan Zhang, Peixian Chen, Yanwei Li, Shaohui Lin, Sirui Zhao, Ke~Li, Tong Xu, Xiawu Zheng, Enhong Chen, Caifeng Shan, Ran He, and Xing Sun.
\newblock Video-mme: The first-ever comprehensive evaluation benchmark of multi-modal llms in video analysis.
\newblock In \emph{Proceedings of the IEEE/CVF Conference on Computer Vision and Pattern Recognition (CVPR)}, pp.\  24108--24118, June 2025{\natexlab{a}}.

\bibitem[Fu et~al.(2025{\natexlab{b}})Fu, Lin, Long, Shen, Dai, Zhao, Zhang, Dong, Li, Wang, Cao, Yin, Ma, Zheng, Ji, Wu, He, Shan, and Sun]{vita}
Chaoyou Fu, Haojia Lin, Zuwei Long, Yunhang Shen, Yuhang Dai, Meng Zhao, Yi-Fan Zhang, Shaoqi Dong, Yangze Li, Xiong Wang, Haoyu Cao, Di~Yin, Long Ma, Xiawu Zheng, Rongrong Ji, Yunsheng Wu, Ran He, Caifeng Shan, and Xing Sun.
\newblock Vita: Towards open-source interactive omni multimodal llm, 2025{\natexlab{b}}.
\newblock URL \url{https://arxiv.org/abs/2408.05211}.

\bibitem[Gong et~al.(2026)Gong, Wang, Wei, Guo, Zhu, and Chen]{echoing_pixels}
Chao Gong, Depeng Wang, Zhipeng Wei, Ya~Guo, Huijia Zhu, and Jingjing Chen.
\newblock Echoingpixels: Aliasing-resistant joint token reduction for audio-visual {LLM}s.
\newblock In \emph{Forty-third International Conference on Machine Learning}, 2026.
\newblock URL \url{https://openreview.net/forum?id=e8JTAKZYt6}.

\bibitem[Guo et~al.(2026{\natexlab{a}})Guo, Luo, Zhang, Wang, and Wang]{tbd}
Xiaoyang Guo, Guoping Luo, Jusheng Zhang, Keze Wang, and Wenhao Wang.
\newblock Token-budget distillation: Transferring full-token semantics to compressed video vision-language models, 2026{\natexlab{a}}.
\newblock URL \url{https://arxiv.org/abs/2608.28138}.

\bibitem[Guo et~al.(2026{\natexlab{b}})Guo, Yang, Man, Yin, Shi, Karanjai, Gnawali, and Zhang]{macer}
Zhenghui Guo, Yilin Yang, Yuanbin Man, Miao Yin, Weidong Shi, Rabimba Karanjai, Omprakash Gnawali, and Chengming Zhang.
\newblock Allocation before ranking: Decoupled token compression for omnillms, 2026{\natexlab{b}}.
\newblock URL \url{https://arxiv.org/abs/2608.01665}.

\bibitem[Hong et~al.(2026)Hong, Yan, Cai, Jiang, Hu, and Xie]{worldsense}
Jack Hong, Shilin Yan, Jiayin Cai, Xiaolong Jiang, Yao Hu, and Weidi Xie.
\newblock Worldsense: Evaluating real-world omnimodal understanding for multimodal llms.
\newblock In C.~Vondrick, B.~Hariharan, C.~Raffel, L.~Pinto, D.~Yang, and A.~Faust (eds.), \emph{International Conference on Learning Representations}, volume 2026, pp.\  52423--52443, 2026.
\newblock URL \url{https://proceedings.iclr.cc/paper_files/paper/2026/file/55dc32df563a35bf406717fb52c118d5-Paper-Conference.pdf}.

\bibitem[Hu et~al.(2022)Hu, yelong shen, Wallis, Allen-Zhu, Li, Wang, Wang, and Chen]{lora}
Edward~J Hu, yelong shen, Phillip Wallis, Zeyuan Allen-Zhu, Yuanzhi Li, Shean Wang, Lu~Wang, and Weizhu Chen.
\newblock Lo{RA}: Low-rank adaptation of large language models.
\newblock In \emph{International Conference on Learning Representations}, 2022.
\newblock URL \url{https://openreview.net/forum?id=nZeVKeeFYf9}.

\bibitem[Huang et~al.(2026)Huang, Huang, Xu, Gu, Tan, Fu, Shen, Liu, Zhang, Zhang, Hu, Dai, Ge, Chen, Li, Wang, and Zhang]{omni_delta}
Haoyang Huang, Wenjie Huang, Tianqi Xu, Hongyaoxing Gu, Kang Tan, Yikai Fu, Yuhao Shen, Tianyu Liu, Baolin Zhang, Jun Zhang, Xinyi Hu, Jun Dai, Shuang Ge, Lei Chen, Yue Li, Mingchen Wang, and Meng Zhang.
\newblock Omnidelta: Skill-driven budget allocation for token compression in omnillms, 2026.
\newblock URL \url{https://arxiv.org/abs/2607.25669}.

\bibitem[Jung et~al.(2026)Jung, Rho, and Chung]{context_guard}
Chaeyoung Jung, Kyeongha Rho, and Joon~Son Chung.
\newblock Keep what audio cannot say: Context-preserving token pruning for omni-llms, 2026.
\newblock URL \url{https://arxiv.org/abs/2605.11605}.

\bibitem[Lee et~al.(2026)Lee, Kim, Kim, and Hong]{a_pack}
Kyeongyoon Lee, Hongyeob Kim, Youngeun Kim, and Sungeun Hong.
\newblock Deferred audio pruning with local audio-visual dynamics for omni-llms, 2026.
\newblock URL \url{https://arxiv.org/abs/2608.08794}.

\bibitem[Li \& Huang(2026)Li and Huang]{dash}
Bingzhou Li and Tao Huang.
\newblock Dash: Dynamic audio-driven semantic chunking for efficient omnimodal token compression, 2026.
\newblock URL \url{https://arxiv.org/abs/2603.15685}.

\bibitem[Li et~al.(2026{\natexlab{a}})Li, Chen, Ji, Xu, Cui, Li, Zhang, Song, Zhang, He, Liu, Wang, Wang, Tang, Wu, Luo, Pan, Xie, Zhang, Wang, Tian, Wang, Cao, Dai, Wang, Wen, MA, Pan, Chang, Taheri, Xia, Plachouras, Benetos, Li, Zhang, Yang, Peng, wang, Liu, Peng, Zhang, and LIU]{omnivideobench}
Caorui Li, Yu~Chen, Yiyan Ji, Jin Xu, Zhenyu Cui, Shihao Li, Yuanxing Zhang, Zhenghao Song, Dingling Zhang, Ying He, Haoxiang Liu, Yuxuan Wang, Qiufeng Wang, Jiafu Tang, Zhenhe Wu, Jiehui Luo, Zhiyu Pan, Weihao Xie, Chenchen Zhang, Zhaohui Wang, Jiayi Tian, Yanghai Wang, Zhe Cao, Minxin Dai, Ke~Wang, Runzhe Wen, Yinghao MA, Yaning Pan, Sungkyun Chang, Termeh Taheri, Haiwen Xia, Christos Plachouras, Emmanouil Benetos, Yizhi Li, Ge~Zhang, Jian Yang, Tianhao Peng, zili wang, Minghao Liu, Junran Peng, Zhaoxiang Zhang, and JIAHENG LIU.
\newblock Omnivideobench: Towards audio-visual understanding evaluation for omni mllms.
\newblock In C.~Vondrick, B.~Hariharan, C.~Raffel, L.~Pinto, D.~Yang, and A.~Faust (eds.), \emph{International Conference on Learning Representations}, volume 2026, pp.\  138214--138236, 2026{\natexlab{a}}.
\newblock URL \url{https://proceedings.iclr.cc/paper_files/paper/2026/file/df5e4da78dfbbfbfad3fcff0c2ce3bd7-Paper-Conference.pdf}.

\bibitem[Li et~al.(2026{\natexlab{b}})Li, Liang, Tian, Wang, Zhang, Yin, Fu, Torr, and Vasconcelos]{s2vopd}
Yijiang Li, Yijun Liang, Yunjie Tian, Bingyang Wang, Ke~Zhang, Zhenfei Yin, Di~Fu, Philip Torr, and Nuno Vasconcelos.
\newblock Self-supervised visual on-policy distillation, 2026{\natexlab{b}}.
\newblock URL \url{https://arxiv.org/abs/2608.14144}.

\bibitem[Liang et~al.(2026)Liang, Zheng, Wang, and Mao]{spiral}
Tianyu Liang, Xiangxi Zheng, Yilin Wang, and Dongxing Mao.
\newblock Same semantics, different paths: Self-improving alignment for vision-text compression, 2026.
\newblock URL \url{https://arxiv.org/abs/2608.02109}.

\bibitem[Lin(1991)]{jsd}
J.~Lin.
\newblock Divergence measures based on the shannon entropy.
\newblock \emph{IEEE Transactions on Information Theory}, 37\penalty0 (1):\penalty0 145--151, 1991.
\newblock \doi{10.1109/18.61115}.

\bibitem[Loshchilov \& Hutter(2019)Loshchilov and Hutter]{adamw}
Ilya Loshchilov and Frank Hutter.
\newblock Decoupled weight decay regularization.
\newblock In \emph{International Conference on Learning Representations}, 2019.
\newblock URL \url{https://openreview.net/forum?id=Bkg6RiCqY7}.

\bibitem[Park et~al.(2026)Park, Jang, Choi, Lee, Choi, and Jeon]{omni_drop}
Yeo~Jeong Park, Hyemi Jang, Minseo Choi, Jongsun Lee, Jooyoung Choi, and Yongkweon Jeon.
\newblock Omnidrop: Layer-wise token pruning for omni-modal llms via query-guidance, 2026.
\newblock URL \url{https://arxiv.org/abs/2605.14458}.

\bibitem[Shang et~al.(2025)Shang, Cai, Xu, Lee, and Yan]{llava_prumerge}
Yuzhang Shang, Mu~Cai, Bingxin Xu, Yong~Jae Lee, and Yan Yan.
\newblock Llava-prumerge: Adaptive token reduction for efficient large multimodal models.
\newblock In \emph{Proceedings of the IEEE/CVF International Conference on Computer Vision (ICCV)}, pp.\  22857--22867, October 2025.

\bibitem[Shao et~al.(2024)Shao, Wang, Zhu, Xu, Song, Bi, Zhang, Zhang, Li, Wu, and Guo]{grpo}
Zhihong Shao, Peiyi Wang, Qihao Zhu, Runxin Xu, Junxiao Song, Xiao Bi, Haowei Zhang, Mingchuan Zhang, Y.~K. Li, Y.~Wu, and Daya Guo.
\newblock Deepseekmath: Pushing the limits of mathematical reasoning in open language models, 2024.
\newblock URL \url{https://arxiv.org/abs/2402.03300}.

\bibitem[Su et~al.(2026{\natexlab{a}})Su, Luo, Ma, Hu, Jin, and Zheng]{omni_scope}
Jinsen Su, Yongdong Luo, Yuexiao Ma, Yibo Hu, Meiguang Jin, and Xiawu Zheng.
\newblock Omniscope: Modality-decoupled token compression for omnimodal large language models, 2026{\natexlab{a}}.
\newblock URL \url{https://arxiv.org/abs/2607.23193}.

\bibitem[Su et~al.(2026{\natexlab{b}})Su, Shi, Liu, Yu, Min, Zhang, Wang, Wang, Liu, Zhang, Wu, Huo, and Ding]{omnipack}
Wanshun Su, Yang Shi, Feihu Liu, Ziwen Yu, Yan Min, Zhuoran Zhang, Qixun Wang, Haotian Wang, Shixuan Liu, Yuanxing Zhang, Peng Wu, Chengfu Huo, and Liang Ding.
\newblock Omnipack: Unified token compression for efficient omni-modal large language models, 2026{\natexlab{b}}.
\newblock URL \url{https://arxiv.org/abs/2608.03812}.

\bibitem[Sun et~al.(2024)Sun, Yu, Tang, Chen, Tan, Li, Lu, Ma, Wang, and Zhang]{video_salmonn}
Guangzhi Sun, Wenyi Yu, Changli Tang, Xianzhao Chen, Tian Tan, Wei Li, Lu~Lu, Zejun Ma, Yuxuan Wang, and Chao Zhang.
\newblock video-{SALMONN}: Speech-enhanced audio-visual large language models.
\newblock In Ruslan Salakhutdinov, Zico Kolter, Katherine Heller, Adrian Weller, Nuria Oliver, Jonathan Scarlett, and Felix Berkenkamp (eds.), \emph{Proceedings of the 41st International Conference on Machine Learning}, volume 235 of \emph{Proceedings of Machine Learning Research}, pp.\  47198--47217. PMLR, 21--27 Jul 2024.
\newblock URL \url{https://proceedings.mlr.press/v235/sun24l.html}.

\bibitem[Tao et~al.(2026)Tao, Shao, Yu, Wang, Liu, and Wang]{omnizip}
Keda Tao, Kele Shao, Bohan Yu, Weiqiang Wang, Jian Liu, and Huan Wang.
\newblock Omnizip: Audio-guided dynamic token compression for fast omnimodal large language models.
\newblock In \emph{Proceedings of the IEEE/CVF Conference on Computer Vision and Pattern Recognition (CVPR)}, pp.\  17682--17692, June 2026.

\bibitem[Wang et~al.(2026{\natexlab{a}})Wang, Zhao, Liang, Ye, Chen, Huang, and Fu]{clue_opsd}
Kaishen Wang, Dongdi Zhao, Yijun Liang, Dingqiang Ye, Ruibo Chen, Heng Huang, and Di~Fu.
\newblock Where to look matters: On-policy self-distillation for long-video understanding, 2026{\natexlab{a}}.
\newblock URL \url{https://arxiv.org/abs/2608.25356}.

\bibitem[Wang et~al.(2026{\natexlab{b}})Wang, Zhang, Tang, Cheng, and Wei]{nopd}
Shuai Wang, Daoan Zhang, Zhe Tang, Hao Cheng, and Jiaheng Wei.
\newblock Self-boosting vision-language models with noisy student on-policy self-distillation, 2026{\natexlab{b}}.
\newblock URL \url{https://arxiv.org/abs/2607.23125}.

\bibitem[Wang et~al.(2026{\natexlab{c}})Wang, Yuan, Zhai, Li, Shu, Gong, Guo, and Liu]{omnifit}
Zining Wang, Zhihang Yuan, Yingjie Zhai, Wenshuo Li, Han Shu, Ruihao Gong, Jinyang Guo, and Xianglong Liu.
\newblock Omnifit: Bridging modalities via layer-adaptive token compression for omnimodal large language models.
\newblock In \emph{Forty-third International Conference on Machine Learning}, 2026{\natexlab{c}}.
\newblock URL \url{https://openreview.net/forum?id=8RY20mLzup}.

\bibitem[Wen et~al.(2025)Wen, Wang, Zhou, Zhang, Zhang, Gao, Chen, Wang, Li, He, and Zhang]{epic}
Zichen Wen, Shaobo Wang, Yufa Zhou, Junyuan Zhang, Qintong Zhang, Yifeng Gao, Zhaorun Chen, Bin Wang, Weijia Li, Conghui He, and Linfeng Zhang.
\newblock Efficient multi-modal large language models via progressive consistency distillation.
\newblock In D.~Belgrave, C.~Zhang, H.~Lin, R.~Pascanu, P.~Koniusz, M.~Ghassemi, and N.~Chen (eds.), \emph{Advances in Neural Information Processing Systems}, volume 38, Main Conference, pp.\  69726--69753. Curran Associates, Inc., 2025.
\newblock \doi{10.52202/085713-2346}.
\newblock URL \url{https://proceedings.neurips.cc/paper_files/paper/2025/file/6518f9339196e172fa0ceef48a85543a-Paper-Conference.pdf}.

\bibitem[Wu et~al.(2026)Wu, Liu, Wu, Chen, and Shen]{o_marc}
Peiran Wu, Yunze Liu, Chi-Hao Wu, Chen Chen, and Junxiao Shen.
\newblock O-marc: Omni memory-augmented compression distillation for efficient video understanding, 2026.
\newblock URL \url{https://arxiv.org/abs/2605.26584}.

\bibitem[Xin et~al.(2026)Xin, Yang, Zhao, Wang, Rao, Lyu, and Li]{seats}
Zijie Xin, Jie Yang, Ruixiang Zhao, Tianyi Wang, Fengyun Rao, Jing Lyu, and Xirong Li.
\newblock Stage-adaptive token selection for efficient omni-modal llms, 2026.
\newblock URL \url{https://arxiv.org/abs/2605.20035}.

\bibitem[Xu et~al.(2025{\natexlab{a}})Xu, Guo, He, Hu, He, Bai, Chen, Wang, Fan, Dang, Zhang, Wang, Chu, and Lin]{qwen2_5_omni}
Jin Xu, Zhifang Guo, Jinzheng He, Hangrui Hu, Ting He, Shuai Bai, Keqin Chen, Jialin Wang, Yang Fan, Kai Dang, Bin Zhang, Xiong Wang, Yunfei Chu, and Junyang Lin.
\newblock Qwen2.5-omni technical report, 2025{\natexlab{a}}.
\newblock URL \url{https://arxiv.org/abs/2503.20215}.

\bibitem[Xu et~al.(2025{\natexlab{b}})Xu, Guo, Hu, Chu, Wang, He, Wang, Shi, He, Zhu, Lv, Wang, Guo, Wang, Ma, Zhang, Zhang, Hao, Guo, Yang, Zhang, Ma, Wei, Bai, Chen, Liu, Wang, Yang, Liu, Ren, Zheng, Men, Zhou, Yu, Yang, Yu, Zhou, and Lin]{qwen3_omni}
Jin Xu, Zhifang Guo, Hangrui Hu, Yunfei Chu, Xiong Wang, Jinzheng He, Yuxuan Wang, Xian Shi, Ting He, Xinfa Zhu, Yuanjun Lv, Yongqi Wang, Dake Guo, He~Wang, Linhan Ma, Pei Zhang, Xinyu Zhang, Hongkun Hao, Zishan Guo, Baosong Yang, Bin Zhang, Ziyang Ma, Xipin Wei, Shuai Bai, Keqin Chen, Xuejing Liu, Peng Wang, Mingkun Yang, Dayiheng Liu, Xingzhang Ren, Bo~Zheng, Rui Men, Fan Zhou, Bowen Yu, Jianxin Yang, Le~Yu, Jingren Zhou, and Junyang Lin.
\newblock Qwen3-omni technical report, 2025{\natexlab{b}}.
\newblock URL \url{https://arxiv.org/abs/2509.17765}.

\bibitem[Yang et~al.(2026)Yang, Xu, Li, Wang, Zhang, Li, Lou, Feng, and Li]{omni_select}
Morunliu Yang, Ruotao Xu, Le~Li, Yue Wang, Jianxin Zhang, Juntao Li, Yihang Lou, Siwei Feng, and Peifeng Li.
\newblock Omniselect: Dynamic modality-aware token compression for efficient omni-modal large language models, 2026.
\newblock URL \url{https://arxiv.org/abs/2605.18041}.

\bibitem[Yang et~al.(2025{\natexlab{a}})Yang, Yao, Chen, Fu, Bai, Zhao, Sun, Yin, Wei, and Zhou]{humanomni_v2}
Qize Yang, Shimin Yao, Weixuan Chen, Shenghao Fu, Detao Bai, Jiaxing Zhao, Boyuan Sun, Bowen Yin, Xihan Wei, and Jingren Zhou.
\newblock Humanomniv2: From understanding to omni-modal reasoning with context, 2025{\natexlab{a}}.
\newblock URL \url{https://arxiv.org/abs/2506.21277}.

\bibitem[Yang et~al.(2025{\natexlab{b}})Yang, Chen, Tian, Wang, Li, Yu, and Jia]{visionzip}
Senqiao Yang, Yukang Chen, Zhuotao Tian, Chengyao Wang, Jingyao Li, Bei Yu, and Jiaya Jia.
\newblock Visionzip: Longer is better but not necessary in vision language models.
\newblock In \emph{Proceedings of the IEEE/CVF Conference on Computer Vision and Pattern Recognition (CVPR)}, pp.\  19792--19802, June 2025{\natexlab{b}}.

\bibitem[Yang et~al.(2025{\natexlab{c}})Yang, Zhuang, Sun, Tang, Li, Li, Jiang, Li, Ma, and Zhang]{avut}
Yudong Yang, Jimin Zhuang, Guangzhi Sun, Changli Tang, Yixuan Li, Peihan Li, Yifan Jiang, Wei Li, Zejun Ma, and Chao Zhang.
\newblock Audio-centric video understanding benchmark without text shortcut.
\newblock In Christos Christodoulopoulos, Tanmoy Chakraborty, Carolyn Rose, and Violet Peng (eds.), \emph{Proceedings of the 2025 Conference on Empirical Methods in Natural Language Processing}, pp.\  6569--6587, Suzhou, China, November 2025{\natexlab{c}}. Association for Computational Linguistics.
\newblock ISBN 979-8-89176-332-6.
\newblock \doi{10.18653/v1/2025.emnlp-main.333}.
\newblock URL \url{https://aclanthology.org/2025.emnlp-main.333/}.

\bibitem[Ye et~al.(2026)Ye, Yang, Goel, Huang, Wan, Tian, Cheng, Zhu, Su, Lou, Lin, Yang, Ghosh, Liu, Chen, Jahangiri, Dantrey, Xu, Hosseini-Asl, Mohseni~Taheri, Nariyambut~Murali, Liu, Lu, Olabiyi, Wang, Valle, Catanzaro, Tao, Han, Kautz, Yin, and Molchanov]{omnivinci}
Hanrong Ye, Chao-Han~Huck Yang, Arushi Goel, Wei Huang, Zhen Wan, Jinchuan Tian, An-Chieh Cheng, Ligeng Zhu, Yuanhang Su, Yuming Lou, Yong-Xiang Lin, Dong Yang, Sreyan Ghosh, Zhijian Liu, Yukang Chen, Ehsan Jahangiri, Ambrish Dantrey, Daguang Xu, Ehsan Hosseini-Asl, Seyed~Danial Mohseni~Taheri, Vidya Nariyambut~Murali, Sifei Liu, Yao Lu, Oluwatobi Olabiyi, Yu-Chiang~Frank Wang, Rafael Valle, Bryan Catanzaro, Andrew Tao, Song Han, Jan Kautz, Hongxu~(Danny) Yin, and Pavlo Molchanov.
\newblock Omnivinci: Enhancing architecture and data for omni-modal understanding llm.
\newblock In C.~Vondrick, B.~Hariharan, C.~Raffel, L.~Pinto, D.~Yang, and A.~Faust (eds.), \emph{International Conference on Learning Representations}, volume 2026, pp.\  56101--56138, 2026.
\newblock URL \url{https://proceedings.iclr.cc/paper_files/paper/2026/file/5c1863f711c721648387ac2ef745facb-Paper-Conference.pdf}.

\bibitem[Ye et~al.(2025)Ye, Gan, Huang, Ge, and Tang]{voco_llama}
Xubing Ye, Yukang Gan, Xiaoke Huang, Yixiao Ge, and Yansong Tang.
\newblock Voco-llama: Towards vision compression with large language models.
\newblock In \emph{Proceedings of the IEEE/CVF Conference on Computer Vision and Pattern Recognition (CVPR)}, pp.\  29836--29846, June 2025.

\bibitem[Yoo et~al.(2026)Yoo, Jang, Cho, and Chung]{remo}
Suho Yoo, Youngjoon Jang, Hyebin Cho, and Joon~Son Chung.
\newblock Out of sight, still in mind: Token compression for omni-llms, 2026.
\newblock URL \url{https://arxiv.org/abs/2607.21179}.

\bibitem[Yuan et~al.(2026)Yuan, Lou, Yu, Lin, Sun, Han, and Lu]{vision_opd}
Qianhao Yuan, Jie Lou, Xing Yu, Hongyu Lin, Le~Sun, Xianpei Han, and Yaojie Lu.
\newblock Vision-opd: Learning to see fine details for multimodal llms via on-policy self-distillation, 2026.
\newblock URL \url{https://arxiv.org/abs/2605.18740}.

\bibitem[Zhao et~al.(2026)Zhao, Xie, Liu, Huang, Pang, Chen, and Grover]{opsd}
Siyan Zhao, Zhihui Xie, Mengchen Liu, Jing Huang, Guan Pang, Feiyu Chen, and Aditya Grover.
\newblock Self-distilled reasoner: On-policy self-distillation for large language models.
\newblock In \emph{Forty-third International Conference on Machine Learning}, 2026.
\newblock URL \url{https://openreview.net/forum?id=Jpxfof0EaS}.

\bibitem[Zhong et~al.(2026)Zhong, Nie, and Shan]{omni_prune}
Yiming Zhong, Chang Nie, and Caifeng Shan.
\newblock Omni-prune: Query-aware unified token pruning for efficient omnimodal large language models, 2026.
\newblock URL \url{https://arxiv.org/abs/2607.23445}.

\bibitem[Zhou et~al.(2026)Zhou, Wang, Wu, and Jiang]{dailyomni}
Ziwei Zhou, Rui Wang, Zuxuan Wu, and Yu-Gang Jiang.
\newblock Daily-omni: Towards audio-visual reasoning with temporal alignment across modalities, 2026.
\newblock URL \url{https://arxiv.org/abs/2505.17862}.

\bibitem[Zhu et~al.(2026)Zhu, Wang, Wen, Zhang, Zhang, Liu, Chen, Wu, Yang, and Jiang]{rp_opsd}
Qihui Zhu, Yuchen Wang, Zijian Wen, Tao Zhang, Mengjie Zhang, Yang Liu, Shuangwu Chen, Siying Wu, Jian Yang, and Xiaofeng Jiang.
\newblock Rp-opsd: Resolution-privileged on-policy self-distillation for multimodal large language models, 2026.
\newblock URL \url{https://arxiv.org/abs/2607.24447}.

\end{thebibliography}
\bibliographystyle{iclr2027_conference}

\appendix
\section{Additional Experimental Analyses}
\label{app:additional_analyses}

\subsection{Task Conditioning and Supervision}
\label{app:task_conditioning}

Table~\ref{tab:app_supervision_controls} examines how training prompts and
reference supervision affect adaptation. The three reference-free CAFD
variants share the distillation mechanism, media sequence, and per-video
exposure counts. The main configuration, labeled CAFD, uses the question
and candidate options; CAFD (question-only) removes the options, while
CAFD (fixed-generic) replaces the question and options with a shared
instruction. All three use a full-token teacher without reference answers or
rationales. Two additional controls retain the main configuration's student
prompt: CAFD + reference supplies the reference answer and rationale to
the teacher, while compressed-input SFT uses them as student training targets.

All three reference-free variants improve over Unadapted. In particular, the fixed-generic
gains show that CAFD can improve compressed inference using only media
and a shared instruction, without per-example questions, options, reference
answers, or rationales as model inputs. This supports adaptation when
task-specific supervision is unavailable. Compressed-input SFT is strongest
overall. Supplying references to the teacher does not improve average accuracy
over the main CAFD configuration.
Appendix~\ref{app:control_details} provides the per-budget results.

\begin{table}[!ht]
    \caption{\textbf{Task conditioning and target supervision.}
    All adapted models are trained with DivPrune at 5\% retention, and all rows are
    evaluated with DivPrune across five deployment budgets; accuracies (\%)
    are budget averages. ``Reference use'' indicates whether reference
    answers and rationales are supplied as teacher input or used as
    student targets; all variants use the same task input at evaluation.}
    \label{tab:app_supervision_controls}
    \centering
    \appendixtablestyle
    \begin{autowidthtabular}{lcccccc}
        \toprule
        Variant & Adaptation prompt & Reference use & WorldSense & DailyOmni &
        AVUT & Avg. \\
        \midrule
        Unadapted & -- & -- & 42.99 & 54.74 & 58.19 & 51.97 \\
\midrule
CAFD (fixed-generic) & Shared generic instruction & None & \textbf{44.21} & 57.03 & 59.19 & 53.47 \\
CAFD (question-only) & Question & None & 44.19 & \textbf{57.26} & \textbf{59.67} & \textbf{53.70} \\
CAFD & Question + options & None & 44.16 & 57.04 & 59.56 & 53.59 \\
\midrule
CAFD + reference & Question + options & Teacher input & 44.17 & 56.66 & 59.49 & 53.44 \\
Compressed-input SFT & Question + options & Student target & \textbf{44.31} & \textbf{57.78} & \textbf{59.90} & \textbf{54.00} \\
\bottomrule
    \end{autowidthtabular}
\end{table}

\subsection{Video-Disjoint Target-Domain Adaptation}
\label{app:target_domain}

To test whether in-domain media can improve compressed inference without
task-specific annotations, we split WorldSense videos into approximately
70\% for adaptation and 30\% for evaluation using seed 42. The two partitions
contain 1,163 and 499 videos, respectively; the evaluation partition
includes 981 QA instances.
Training reads only paths and IDs from the training partition, cycles them
to 5,700 exposures, and uses the fixed-generic prompt; it reads no question,
candidate option, answer, rationale, category, or evaluation field. All
three compressed conditions in Table~\ref{tab:target_domain_results} are
recomputed on the identical held-out QA identities.
The OmniVideo control uses the same fixed-generic checkpoint as
Table~\ref{tab:app_supervision_controls}. We compute its accuracy on the same
981 held-out QA instances using the existing full-WorldSense predictions.

In-domain adaptation raises average accuracy from 43.08\% to 44.00\%,
compared with 43.73\% for the OmniVideo-trained control. These results show
that in-domain audio-video data and a shared generic instruction can improve
compressed inference on unseen videos without task-specific annotations.

\begin{table}[h]
    \centering
    \appendixtablestyle
    \caption{WorldSense held-out accuracy (\%) under video-disjoint
    fixed-generic adaptation. Both models are trained with DivPrune at
    5\% nominal retention; compressed evaluations use DivPrune at the five
    listed budgets. The full-token reference is reported in the
    Avg. column as a single full-input accuracy.}
    \label{tab:target_domain_results}
    \begin{autowidthtabular}{lrrrrrr}
        \toprule
        Model & 35\% & 25\% & 15\% & 10\% & 5\% & Avg. \\
        \midrule
        Full-token reference & -- & -- & -- & -- & -- & 46.99 \\
\midrule
Unadapted & 46.59 & 45.57 & 43.32 & 41.59 & 38.33 & 43.08 \\
OmniVideo fixed-generic & 46.08 & 45.67 & 44.14 & 43.83 & 38.94 & 43.73 \\
WorldSense video-disjoint fixed-generic & 46.28 & 45.77 & 44.75 & 43.63 & 39.55 & 44.00 \\
\bottomrule
    \end{autowidthtabular}
\end{table}

\subsection{Training-Seed Sensitivity}
\label{app:seed_sensitivity}

To assess sensitivity to training randomness, we repeat DivPrune adaptation
at 5\% retention with three seeds (42, 142, and 242). Averaged across three
benchmarks and five deployment budgets, the runs achieve 53.59\%, 53.65\%,
and 53.47\% accuracy, respectively, yielding \(53.57\pm0.09\%\)
(mean $\pm$ sample standard deviation). The small variation indicates that
the reported performance is consistent across training seeds.
Table~\ref{tab:seed_details} provides the per-budget results.

\begin{table}[h]
    \centering
    \appendixtablestyle
    \caption{Per-budget accuracy (\%) for three independent DivPrune
    training seeds. All models are trained with DivPrune at 5\% nominal
    retention and evaluated with DivPrune at the five listed budgets.
    Column Avg. averages deployment budgets; block-ending
    Avg. averages benchmarks, and their intersection averages all 15 conditions.}
    \label{tab:seed_details}
    \begin{autowidthtabular}{llrrrrrr}
        \toprule
        Seed & Benchmark & 35\% & 25\% & 15\% & 10\% & 5\% & Avg. \\
        \midrule
        42 & WorldSense & 46.37 & 46.15 & 44.45 & 43.35 & 40.45 & 44.16 \\
 & DailyOmni & 60.82 & 58.73 & 58.06 & 55.22 & 52.38 & 57.04 \\
 & AVUT & 63.21 & 62.00 & 60.03 & 58.30 & 54.27 & 59.56 \\
\cmidrule(l){2-8}
 & Avg. & 56.80 & 55.63 & 54.18 & 52.29 & 49.03 & 53.59 \\
\midrule
142 & WorldSense & 46.41 & 46.06 & 44.80 & 43.44 & 40.29 & 44.20 \\
 & DailyOmni & 60.57 & 59.15 & 57.98 & 55.47 & 52.13 & 57.06 \\
 & AVUT & 63.32 & 62.05 & 60.03 & 58.59 & 54.50 & 59.70 \\
\cmidrule(l){2-8}
 & Avg. & 56.77 & 55.75 & 54.27 & 52.50 & 48.97 & 53.65 \\
\midrule
242 & WorldSense & 46.25 & 46.19 & 44.58 & 43.22 & 40.38 & 44.12 \\
 & DailyOmni & 60.32 & 58.81 & 58.23 & 55.22 & 51.80 & 56.88 \\
 & AVUT & 63.21 & 61.59 & 59.98 & 58.13 & 54.15 & 59.41 \\
\cmidrule(l){2-8}
 & Avg. & 56.59 & 55.53 & 54.26 & 52.19 & 48.78 & 53.47 \\
\bottomrule
    \end{autowidthtabular}
\end{table}

\section{Reproducibility Details}
\label{app:reproducibility}

\subsection{Datasets and Statistics}
\label{app:benchmarks}

Table~\ref{tab:benchmark_statistics} summarizes the local training subset and
the five evaluation sets. Video counts and duration statistics are computed
once per distinct video, whereas QA counts reflect the examples used for
training or evaluation. Durations describe the media files before frame
sampling and audio truncation.

\begin{table}[htbp]
    \caption{\textbf{Dataset statistics for the sets used in this study.}
    Durations are in seconds and are computed per distinct video rather than
    per QA example. The OmniVideo row is the selected training subset, not
    the full OmniVideo-100K dataset; the AVUT row is its human-annotated subset.}
    \label{tab:benchmark_statistics}
    \centering
    \appendixtablestyle
    \begin{autowidthtabular}{lrrrrl}
        \toprule
        Dataset & Videos & QAs & Mean & Median & Min--max \\
        \midrule
        OmniVideo train & 1,485 & 5,700 & 103.99 & 100.00 & 53.83--179.77 \\
        WorldSense & 1,662 & 3,172 & 141.14 & 89.68 & 15.60--656.59 \\
        DailyOmni & 684 & 1,197 & 42.83 & 30.03 & 8.96--60.03 \\
        AVUT & 691 & 1,734 & 69.03 & 58.76 & 5.30--919.00 \\
        OmniVideoBench & 628 & 1,000 & 385.53 & 280.92 & 4.92--1,955.77 \\
        Video-MME & 900 & 2,700 & 1,021.27 & 487.94 & 11.03--3,579.45 \\
        \bottomrule
    \end{autowidthtabular}
\end{table}

\paragraph{Training data.}
OmniVideo-100K~\citep{omnivideo_100k} provides audio-visual questions built
from structured video descriptions and evidence chains. We use a 5,700-QA
subset spanning ten task types and 1,485 videos. The main adaptation prompt
uses its questions and options, but neither its prescribed answers nor its
reference rationales are optimization targets; the fixed-generic variant uses
the media with a shared prompt but no per-example QA fields.

\paragraph{Evaluation sets.}
WorldSense~\citep{worldsense} tests joint audio-visual understanding of
real-world videos across diverse domains and tasks. DailyOmni~\citep{dailyomni}
emphasizes temporal alignment between audible and visible events in everyday
videos. AVUT~\citep{avut} focuses on audio-centric video understanding and
audio-visual interactions; we evaluate its human-annotated subset.
OmniVideoBench~\citep{omnivideobench} evaluates audio-visual reasoning over
videos with varied content and duration. Video-MME~\citep{video_mme} covers
short, medium, and long videos; we use its video-and-audio setting without
subtitles. All five sets are evaluated as multiple-choice QA under the common
protocol in Section~\ref{sec:experimental_setup}.

\subsection{Data Preparation}
\label{app:media_preparation}
\label{app:data_protocol}

\paragraph{Video transcoding.}
To ensure compatibility with our Decord-based evaluation pipeline, we
converted 397 VP9-encoded AVUT videos to H.264 and retained the remaining
294 H.264 videos without re-encoding. All 691 prepared videos decoded
successfully, with audio, resolution, and duration preserved.
In OmniVideoBench, one VP9 file that failed with the same reader was
converted to H.264; its audio was copied, and its resolution, frame count,
and duration were unchanged. The QA annotations and evaluation splits were
not modified by these media conversions.

\paragraph{Training data construction.}

We construct our training set from OmniVideo-100K's 29,966-example MCQ
branch, excluding questions with duplicate options or placeholder-only text
and two-option event-ordering questions. With seed 42, we partition the
videos and select 6,000 QA pairs with balanced task quotas: 5,700 for
training and 300 for validation, with no shared videos between the splits.
The training set contains 5,700 QA pairs from 1,485 videos, balanced across
ten task types with 570 QA pairs per task. Each video contributes three or
four QA pairs. The reserved validation split is not
used for model evaluation or selection.

Both the fixed-generic and question-only
variants preserve the main configuration's media sequence and per-video
exposure counts. The fixed-generic variant uses a media-only projection of
the same 5,700-row QA-derived manifest, retaining only IDs and media paths,
and applies a shared instruction without reading per-example QA fields.
The question-only variant retains each question but removes its candidate
options. These controls keep media sampling fixed while changing the
information included in the training prompt.

We checked for overlap between our training set and the five evaluation
benchmarks using video identifiers, normalized
questions, file hashes where possible, and sampled-frame fingerprints. No
overlap was found with WorldSense, DailyOmni, AVUT, OmniVideoBench, or
Video-MME under these tests.

\subsection{Compressor Implementations}
\label{app:compression_details}

Global Uniform applies segment-midpoint sampling to the complete interleaved
audio-video token sequence. DivPrune forms separate audio and video groups
and globally retains diverse tokens within each modality. DivPrune-Merge
uses the identical anchors and token budget, assigns every discarded token
to its nearest same-modality anchor, and replaces anchors that receive
discarded tokens by
\[
    0.7\,\mathbf{h}_{\mathrm{anchor}}
    +0.3\,\operatorname{mean}_{i\mapsto\mathrm{anchor}}\mathbf{h}_i.
\]
OmniZip's separate audio/video ratio controls do not consistently target a
low joint budget when the modality mix varies across samples. OmniZip*
therefore replaces only its budget controller with a capacity-aware,
audio-favored joint allocator; its importance scoring, audio merging, and ISTM
operations are unchanged. We use an audio-favored multiplier of 3, a
contextual-to-video ratio of 0.2, and \(g=3\).
SEATS prunes audio-video tokens both before and within the LLM, so we report
its layer-average retention estimate rather than treating the final-layer token count
as overall retention.

\subsection{Compression-Budget Configurations}
\label{app:budget_configurations}

Table~\ref{tab:train_retention_mapping} summarizes the compression-budget
configurations shared by training and evaluation. For each compressor and
nominal budget, the listed parameters are fixed across all six datasets
(OmniVideo and the five evaluation benchmarks) wherever that setting is used.
Uniform uses a joint audio-video retention ratio.
DivPrune and DivPrune-Merge share base audio/video allocation ratios, which
are rescaled per sample to the joint budget in the first column.
OmniZip* specifies a joint request, while SEATS specifies layer-average
audio/video targets. These are configuration parameters; benchmark-specific
evaluation retention and computation are reported separately in
Table~\ref{tab:retention_flops}.

\begin{table}[htbp]
    \centering
    \appendixtablestyle
    \caption{\textbf{Compression-budget configurations for five compression pipelines.}
    All values are percentages. Table~\ref{tab:main_results} uses the 5\% row
    for training and the 5\%, 10\%, 15\%, 25\%, and 35\% rows for evaluation.
    The seven training budgets for DivPrune, OmniZip*, and SEATS are used in
    Figure~\ref{fig:train_rtt_and_datacount}(b).
    Superscript $*$ marks configurations provided for completeness but not
    used for training in our experiments. DivPrune-M denotes DivPrune-Merge.}
    \label{tab:train_retention_mapping}
    \begin{autowidthtabular}{lccccc}
        \toprule
        Nominal budget & Uniform & DivPrune & DivPrune-M & OmniZip* & SEATS \\
         & Joint & Base audio/video & Base audio/video & Joint request & Layer-avg. audio/video \\
        \midrule
        35\% & $35^{*}$ & 65/30 & $65/30^{*}$ & 35 & 65/30 \\
        25\% & $25^{*}$ & 55/20 & $55/20^{*}$ & 25 & 55/20 \\
        15\% & $15^{*}$ & 45/10 & $45/10^{*}$ & 15 & 45/10 \\
        10\% & $10^{*}$ & 35/6 & $35/6^{*}$ & 10 & 35/6 \\
        5\% & 5 & 17.5/3 & 17.5/3 & 5 & 9.4353/3.1451 \\
        3\% & $3^{*}$ & 10.5/1.8 & $10.5/1.8^{*}$ & 3 & 5.66118/1.88706 \\
        1\% & $1^{*}$ & 3.5/0.6 & $3.5/0.6^{*}$ & 1.2 & 1.88706/0.62902 \\
        \bottomrule
    \end{autowidthtabular}
\end{table}

\subsection{Training Prompts}
\label{app:prompts}

All three reference-free CAFD variants use explicit system messages with the same
Qwen identity description but different task instructions: describing
audiovisual content, answering the question, or selecting an MCQ option.
The boxes below show their training user messages and the teacher's user
message for CAFD + reference, with line wrapping changed only for typesetting.

\noindent\begin{minipage}{\linewidth}
\paragraph{CAFD (fixed-generic).}
\begin{promptbox}[listparameters={\vspace{10pt}}]
Describe the important auditory and visual content of this video.
Identify the main entities, events, actions, sounds,
speech, and their temporal relationships.

Use exactly the following output format:

<analysis>
Your concise evidence-based analysis here.
</analysis>
<answer>Your concise overall description here.</answer>

Do not output anything after </answer>.
\end{promptbox}
\end{minipage}

\paragraph{CAFD (question-only).}
\begin{promptbox}
Question:
{question}

First provide a concise, evidence-based analysis
using the relevant visual and audio information.
Then give a concise final answer to the question.

Use exactly the following output format:

<analysis>
Your concise reasoning here.
</analysis>
<answer>Your concise answer here.</answer>

Answer the question directly rather than with a letter label.
Do not output anything after </answer>.
\end{promptbox}

\paragraph{CAFD.}
\begin{promptbox}
Question:
{question}

Options:
{label_1}. {option_1}
...
{label_k}. {option_k}

Valid option labels: {label_1}, ..., {label_k}.

First provide a concise, evidence-based analysis
using the relevant visual and audio information.
Then give exactly one final option label.

Use exactly the following output format:

<analysis>
Your concise reasoning here.
</analysis>
<answer>X</answer>

X must be exactly one of the valid option labels listed above.
Do not include the option text or any additional words
inside <answer>...</answer>.
Do not output anything after </answer>.
\end{promptbox}

\paragraph{CAFD + reference.}
The student uses the main CAFD template. The teacher uses the
same system message and the following user message:
\begin{promptbox}
Question:
{question}

Options:
{label_1}. {option_1}
...
{label_k}. {option_k}

The following is authoritative privileged reference information
for this training example. Use it silently to solve the original
question. Do not mention the existence of this reference and do
not mechanically copy its wording.

<reference_analysis>
{analysis_connections}
</reference_analysis>

<reference_answer>
{answer_label}. {answer_text}
</reference_answer>

Valid option labels: {label_1}, ..., {label_k}.

First provide a concise, evidence-based analysis
using the relevant visual and audio information.
Then give exactly one final option label.

Use exactly the following output format:

<analysis>
Your concise reasoning here.
</analysis>
<answer>X</answer>

X must be exactly one of the valid option labels listed above.
Do not include the option text or any additional words
inside <answer>...</answer>.
Do not output anything after </answer>.
\end{promptbox}

In all three reference-free CAFD variants, both paths receive the same
textual prompt; only their media-token views differ. CAFD + reference
additionally inserts the reference rationale and answer into a teacher-only block while
retaining on-policy distillation. Compressed-input SFT uses the same system
and user messages as the main CAFD configuration. Its supervision target follows
the requested output format, with the reference rationale inside
\texttt{<analysis>} and the correct option label inside \texttt{<answer>};
cross-entropy is applied to this reference response.

\subsection{Training Configuration}
\label{app:training_configuration}

Table~\ref{tab:training_configuration} summarizes the main training recipe.
We train only the LoRA parameters in the Thinker decoder, keeping the
original model weights and compression configuration fixed. In this
implementation, the trainable parameters \(\theta\) in Section~\ref{sec:method}
are the LoRA parameters; the teacher EMA is applied to these parameters.
We merge the learned LoRA weights into the LLM before evaluation, so no
separate adapter module is used at inference. Training-data construction,
prompt templates, and compressor-specific settings are given in
Appendices~\ref{app:data_protocol}, \ref{app:prompts}, and
\ref{app:budget_configurations}, respectively.

In the initial development experiments, we selected the 5\% training setting
by average accuracy across WorldSense,
DailyOmni, AVUT, and all five evaluation budgets
with DivPrune, OmniZip*, and SEATS; OmniVideoBench and Video-MME were not
used for this selection.

\begingroup
\appendixlongtablestyle
\begin{longtable}{@{}p{0.32\linewidth}p{0.65\linewidth}@{}}
\caption{Training configuration for the main CAFD runs.}
\label{tab:training_configuration}\\
\toprule
Configuration & Setting \\
\midrule
\endfirsthead
\multicolumn{2}{c}{Table~\thetable\ continued}\\
\toprule
Configuration & Setting \\
\midrule
\endhead
\bottomrule
\endfoot
\textbf{Model and training data} & \\
Backbone & Qwen2.5-Omni-7B \\
Training set & 5,700 QA instances from 1,485 OmniVideo videos \\
Training retention & 5\% nominal budget, fixed for each compressor \\
Video sampling & 2 FPS; at most 128 frames \\
Per-frame pixel bounds & Minimum = maximum = 100,352 \\
LoRA scope & Attention and MLP projections in all 28 Thinker decoder layers \\
LoRA rank / $\alpha$ / dropout & 64 / 128 / 0; no trainable bias \\
\midrule
\textbf{Optimization} & \\
Optimizer & AdamW~\citep{adamw}; $\beta_1=0.9$, $\beta_2=0.999$, $\epsilon=10^{-8}$ \\
Learning rate / weight decay & $5\times10^{-6}$ / 0 \\
Schedule & 5\% linear warmup (18 updates), then linear decay \\
Hardware & 8 NVIDIA RTX 6000D GPUs \\
Batch size & 1 per GPU; 4 accumulation steps; 32 globally \\
Training duration & 2 epochs; 358 optimizer updates \\
Random seed & 42 \\
Gradient clipping & Maximum gradient norm 1 \\
Checkpoint selection & Final Student checkpoint \\
\midrule
\textbf{Rollout and distillation} & \\
Sampling & Temperature 1; top-$p=1$; top-$k=20$ \\
Generation limit / stop & 512 new tokens / \texttt{</answer>} \\
Teacher EMA decay & 0.999 \\
Distillation objective & Symmetric JSD, $\beta=0.5$, temperature 1 \\
Pointwise clipping & Upper bound 0.05 on each vocabulary-level contribution \\
\midrule
\textbf{Implementation} & \\
Precision / attention & bfloat16 / FlashAttention-2~\citep{flashattn2} \\
Distributed training & DeepSpeed ZeRO-2, without parameter or optimizer offload \\
Gradient checkpointing & Enabled except for SEATS \\
\end{longtable}
\endgroup

SEATS disables gradient checkpointing because its gradient-connected prompt
KV-cache path is incompatible with checkpoint recomputation.

CAFD + reference and compressed-input SFT match the main run's training data, student
input, optimizer settings, and update count. SFT retains complete reference
responses with a 1,536-token length check, whereas distillation rollouts
are capped at 512 tokens; supervision-token counts and training FLOPs are
not matched.

\subsection{Evaluation Protocol}
\label{app:evaluation_protocol}

\paragraph{Inputs and preprocessing.}
The default evaluation uses all QA examples in the five evaluation sets
listed in Table~\ref{tab:evaluation_configuration}. Videos are sampled at
a target rate of 2 FPS subject to the benchmark-specific frame cap, with
per-frame minimum and maximum pixel bounds both set to 100,352. We use
available audio tracks without subtitles. WorldSense, DailyOmni, and AVUT
use the full available audio; OmniVideoBench and Video-MME use only the
first 300 seconds of audio. This audio limit does not truncate the video.

\begin{table}[htbp]
    \caption{Default evaluation inputs. The pixel bounds are 100,352 for
    all benchmarks; subtitles are not used.}
    \label{tab:evaluation_configuration}
    \centering
    \appendixtablestyle
    \begin{autowidthtabular}{lrrrl}
        \toprule
        Benchmark & QAs & Target FPS & Frame cap & Audio \\
        \midrule
        WorldSense & 3,172 & 2 & 128 & Full available track \\
        DailyOmni & 1,197 & 2 & 128 & Full available track \\
        AVUT & 1,734 & 2 & 128 & Full available track \\
        OmniVideoBench & 1,000 & 2 & 256 & First 300 seconds \\
        Video-MME & 2,700 & 2 & 768 & First 300 seconds \\
        \bottomrule
    \end{autowidthtabular}
\end{table}

\paragraph{Evaluation prompts.}
Unadapted models and adapted variants use the same MCQ system and user
messages, including all five training variants in
Table~\ref{tab:app_supervision_controls}. The user message contains the video
and the text shown below. Candidate options retain their labels and are
listed in the order supplied by each question; the number of options
depends on the question. Evaluation requests the option letter directly,
without the analysis-plus-answer format used during training. Line wrapping
in the following templates is adjusted only for typesetting.

\paragraph{System message.}
\begin{promptbox}
You are Qwen, a virtual human developed by the Qwen Team,
Alibaba Group, capable of perceiving auditory and visual inputs,
as well as generating text and speech. Please analyze the video
carefully and select the most appropriate answer from the given
options.
\end{promptbox}

\paragraph{User message.}
\begin{promptbox}
{question}
Options:
{candidates_text}
Answer with the option's letter from the given choices directly.
\end{promptbox}
Here, \texttt{\{candidates\_text\}} lists one labeled option per line.

\paragraph{Decoding and scoring.}
We use greedy decoding with a single beam, no sampling, and at most 2 new
text tokens; audio generation is disabled. After removing special tokens
and stripping leading and trailing whitespace, we read the initial option
letter case-insensitively and compare it with the correct label. Empty
responses and responses beginning with an invalid option are scored as
incorrect. Accuracy is the number of correct predictions divided by the
number of evaluated QA examples, reported as a percentage.

\subsection{Comparison Setup}
\label{app:comparison_setup}

\paragraph{Models and budgets.}
The main comparison pairs the original model (\emph{Unadapted}) with its
compressor-matched adapted model (\emph{+CAFD}); \emph{Full tokens}
denotes the original model evaluated without compression. For each of the
five compressors, one model is trained at the nominal 5\% retention
setting and evaluated with that compressor at 35\%, 25\%, 15\%, 10\%, and
5\%, without retraining for each evaluation budget. Each matched pair uses
the same inputs and evaluation protocol and differs only in Thinker weights.

For a given compressor and nominal budget, compression hyperparameters
remain fixed across all five benchmarks and before and after adaptation.
The budget labels denote nominal retention settings; realized retention
can differ across compressors and benchmarks. Appendix~\ref{app:budget_configurations}
lists the settings, and Table~\ref{tab:retention_flops} gives the paired
retention statistics and analytical prefill computation, including the
layer-average reporting convention for SEATS.

\paragraph{Aggregation.}
Each benchmark accuracy weights its QA examples equally. A five-budget
summary is the arithmetic average of that benchmark's accuracies at the
five evaluation budgets. Cross-benchmark averages weight benchmarks
equally: three-benchmark summaries use WorldSense, DailyOmni, and AVUT;
five-benchmark summaries additionally include OmniVideoBench and Video-MME.
Thus, each compressor's overall average in Table~\ref{tab:main_results}
equally weights its 25 benchmark--budget conditions, rather than pooling
QA examples across benchmarks.
All averages and differences are computed from unrounded values and rounded
only for reporting.

For Figure~\ref{fig:aggregate_results}(b), each compressor's accuracy-gap
recovery is its mean adaptation gain divided by the gap between the five-benchmark
full-token mean and its Unadapted mean over 25 benchmark--budget conditions,
multiplied by 100. Mean accuracy-gap recovery averages these five compressor-level
percentages equally, using unrounded inputs throughout.

\section{Detailed Experimental Results}
\label{app:detailed_results}

\subsection{Detailed Compression Results}
\label{app:compression_results}

Table~\ref{tab:retention_flops} expands Table~\ref{tab:main_results} with
retention and analytical prefill computation for each evaluation condition.
For each compressor and nominal budget, compression hyperparameters are
fixed across the five benchmarks and both adaptation states.
Pre-LLM retention is measured over the corpus as the total number of retained
tokens divided by the total number of original tokens for the corresponding
modality or their combination. Nominal budget labels need not equal the
resulting total retention.
Since adaptation updates only the LLM, the pre-LLM compressors operate on
unchanged encoder features. Under the fixed configurations, all 100 pre-LLM comparisons have identical
token counts and analytical prefill computation before and after adaptation.
For SEATS, the reported retention estimates are unchanged, while 20 of 25
comparisons exhibit small changes in decoder-layer computation; all paired
compute values agree at the two-decimal precision shown in the table.
SEATS performs model-dependent pruning within the LLM, which can change
intermediate-layer token counts after adaptation even with fixed
hyperparameters. The unchanged SEATS retention
estimates therefore do not establish identical realized retention at every
layer.

Following the LLM-prefill compute accounting used by SEATS~\citep{seats},
as clarified by its authors, we use the same dense matrix-multiplication
proxy for all five compressors and the full-token reference, counting one
multiply--accumulate as one unit:
\begin{align}
    f(N) &= 2Nd^2 + 2Ndd_{\mathrm{kv}} + 2N^2d + 3Ndd_{\mathrm{ff}},
    \label{eq:prefill_layer_compute}\\
    C_i &= \sum_{\ell=1}^{L} f(N_{i,\ell}) + N_{i,\mathrm{head}}dV,
    \qquad
    \overline C_{\mathrm{T}} = \frac{1}{M\,10^{12}}\sum_{i=1}^{M} C_i.
    \label{eq:prefill_average_compute}
\end{align}
Here \(d\), \(d_{\mathrm{ff}}\), \(V\), and \(L\) denote the decoder
hidden width, FFN intermediate width, vocabulary size, and number of layers.
The output width of each key or value projection is
\(d_{\mathrm{kv}}=H_{\mathrm{kv}}d_h\),
where \(H_{\mathrm{kv}}\) is the number of key/value heads and \(d_h\) is
the head dimension. The terms in \(f(N)\) count the Q/O projections,
K/V projections, attention matrix multiplications, and gated FFN projections,
respectively; the final term in \(C_i\) accounts for the LM head.
Thus, \(f(N)\) is the estimated cost of one decoder layer with \(N\)
input tokens, \(C_i\) is the total prefill cost for QA instance \(i\),
and \(\overline C_{\mathrm{T}}\) is its dataset mean in units of \(10^{12}\).

For QA instance \(i\), \(N_{i,\ell}\) is the sequence length entering
decoder layer \(\ell\), including retained audio-video tokens, text, and
special tokens; \(N_{i,\mathrm{head}}\) is the length passed to the LM head.
Full tokens and the four pre-LLM compressors use a constant length \(N_i\)
across decoder layers and at the LM head, so their cost simplifies to
\(Lf(N_i)+N_i dV\).
SEATS uses each layer's input length in its runtime counter: pruning occurs
after a complete decoder layer and reduces the input length of subsequent
layers. These lengths come from the model's runtime sequences, not from
multiplying the original length by the nominal retention budget.
We average over all \(M\) evaluated QA instances and round only for display.
This proxy counts the listed matrix multiplications, excluding modality
encoders, compression operations, and subsequent autoregressive decoding;
the attention term is not adjusted for causal sparsity or kernel implementation.

\begingroup
\appendixlongtablestyle
\begin{longtable}{lrrrrrr}
\caption{\textbf{Detailed results before and after adaptation.}
Paired entries report Unadapted / +CAFD. All adapted models are trained at
5\% nominal retention and evaluated with the same compressor.
Retention is measured for pre-LLM methods; SEATS reports modality-wise
layer-average targets and an input-token-weighted total estimate.
Full tokens denotes the unadapted full-token reference;
DivPrune-M denotes DivPrune-Merge.}
\label{tab:retention_flops}\\
\toprule
Compressor & Budget & Audio & Video & Total & Prefill compute & Acc. (\%) \\
 & & \multicolumn{3}{c}{Retention (\%)} & (T) & \\
\midrule
\endfirsthead
\multicolumn{7}{c}{\tablename\ \thetable\ continued}\\
\toprule
Compressor & Budget & Audio & Video & Total & Prefill compute & Acc. (\%) \\
 & & \multicolumn{3}{c}{Retention (\%)} & (T) & \\
\midrule
\endhead
\midrule
\multicolumn{7}{r}{Continued on next page}\\
\endfoot
\bottomrule
\endlastfoot
\multicolumn{7}{l}{\textbf{WorldSense}} \\*
Full tokens & 100\% & 100.00 & 100.00 & 100.00 & 117.80 & 46.85 \\*
\midrule
Uniform & 35\% & 34.99 / 34.99 & 35.00 / 35.00 & 35.00 / 35.00 & 34.58 / 34.58 & 43.28 / \textbf{44.96} \\
Uniform & 25\% & 25.00 / 25.00 & 25.00 / 25.00 & 25.00 / 25.00 & 24.17 / 24.17 & 41.96 / \textbf{43.66} \\
Uniform & 15\% & 15.00 / 15.00 & 15.00 / 15.00 & 15.00 / 15.00 & 14.40 / 14.40 & 38.84 / \textbf{39.97} \\
Uniform & 10\% & 10.00 / 10.00 & 10.00 / 10.00 & 10.00 / 10.00 & 9.75 / 9.75 & 36.44 / \textbf{37.64} \\
Uniform & 5\% & 5.02 / 5.02 & 4.99 / 4.99 & 5.00 / 5.00 & 5.26 / 5.26 & 35.18 / \textbf{36.66} \\
\midrule
DivPrune & 35\% & 53.82 / 53.82 & 27.13 / 27.13 & 35.00 / 35.00 & 34.58 / 34.58 & 45.49 / \textbf{46.37} \\
DivPrune & 25\% & 42.91 / 42.91 & 17.51 / 17.51 & 25.00 / 25.00 & 24.17 / 24.17 & 45.02 / \textbf{46.15} \\
DivPrune & 15\% & 31.24 / 31.24 & 8.21 / 8.21 & 15.00 / 15.00 & 14.40 / 14.40 & 43.25 / \textbf{44.45} \\
DivPrune & 10\% & 22.66 / 22.66 & 4.70 / 4.70 & 10.00 / 10.00 & 9.75 / 9.75 & 42.06 / \textbf{43.35} \\
DivPrune & 5\% & 11.33 / 11.33 & 2.35 / 2.35 & 5.00 / 5.00 & 5.26 / 5.26 & 39.12 / \textbf{40.45} \\
\midrule
DivPrune-M & 35\% & 53.82 / 53.82 & 27.13 / 27.13 & 35.00 / 35.00 & 34.58 / 34.58 & 45.40 / \textbf{45.68} \\
DivPrune-M & 25\% & 42.91 / 42.91 & 17.51 / 17.51 & 25.00 / 25.00 & 24.17 / 24.17 & 44.45 / \textbf{45.81} \\
DivPrune-M & 15\% & 31.24 / 31.24 & 8.21 / 8.21 & 15.00 / 15.00 & 14.40 / 14.40 & 43.69 / \textbf{44.99} \\
DivPrune-M & 10\% & 22.66 / 22.66 & 4.70 / 4.70 & 10.00 / 10.00 & 9.75 / 9.75 & 42.72 / \textbf{44.04} \\
DivPrune-M & 5\% & 11.33 / 11.33 & 2.35 / 2.35 & 5.00 / 5.00 & 5.26 / 5.26 & 39.47 / \textbf{41.05} \\
\midrule
OmniZip* & 35\% & 62.37 / 62.37 & 26.16 / 26.16 & 36.84 / 36.84 & 36.52 / 36.52 & 44.04 / \textbf{45.71} \\
OmniZip* & 25\% & 44.55 / 44.55 & 18.35 / 18.35 & 26.07 / 26.07 & 25.24 / 25.24 & 43.25 / \textbf{44.58} \\
OmniZip* & 15\% & 26.73 / 26.73 & 10.73 / 10.73 & 15.45 / 15.45 & 14.82 / 14.82 & 39.97 / \textbf{41.24} \\
OmniZip* & 10\% & 17.82 / 17.82 & 6.95 / 6.95 & 10.15 / 10.15 & 9.89 / 9.89 & 38.75 / \textbf{40.64} \\
OmniZip* & 5\% & 8.91 / 8.91 & 3.21 / 3.21 & 4.89 / 4.89 & 5.17 / 5.17 & 34.93 / \textbf{36.48} \\
\midrule
SEATS & 35\% & 65.00 / 65.00 & 30.00 / 30.00 & 40.32 / 40.32 & 39.94 / 39.94 & 46.44 / \textbf{47.51} \\
SEATS & 25\% & 55.00 / 55.00 & 20.00 / 20.00 & 30.32 / 30.32 & 29.02 / 29.02 & 45.84 / \textbf{47.32} \\
SEATS & 15\% & 45.00 / 45.00 & 10.00 / 10.00 & 20.32 / 20.32 & 19.23 / 19.23 & 44.58 / \textbf{45.62} \\
SEATS & 10\% & 35.00 / 35.00 & 6.00 / 6.00 & 14.55 / 14.55 & 13.30 / 13.30 & 43.32 / \textbf{44.10} \\
SEATS & 5\% & 9.44 / 9.44 & 3.15 / 3.15 & 5.00 / 5.00 & 5.16 / 5.16 & 37.89 / \textbf{39.31} \\
\midrule
\multicolumn{7}{l}{\textbf{DailyOmni}} \\*
Full tokens & 100\% & 100.00 & 100.00 & 100.00 & 65.04 & 62.91 \\*
\midrule
Uniform & 35\% & 34.97 / 34.97 & 35.00 / 35.00 & 35.00 / 35.00 & 20.70 / 20.70 & 57.23 / \textbf{58.65} \\
Uniform & 25\% & 25.03 / 25.03 & 25.00 / 25.00 & 25.00 / 25.00 & 14.79 / 14.79 & 52.05 / \textbf{55.89} \\
Uniform & 15\% & 14.90 / 14.90 & 15.01 / 15.01 & 15.00 / 15.00 & 9.11 / 9.11 & 49.37 / \textbf{52.72} \\
Uniform & 10\% & 10.00 / 10.00 & 10.00 / 10.00 & 10.00 / 10.00 & 6.36 / 6.36 & 45.45 / \textbf{48.45} \\
Uniform & 5\% & 5.03 / 5.03 & 5.00 / 5.00 & 5.00 / 5.00 & 3.68 / 3.68 & 40.77 / \textbf{43.69} \\
\midrule
DivPrune & 35\% & 64.64 / 64.64 & 29.84 / 29.84 & 35.00 / 35.00 & 20.70 / 20.70 & 58.48 / \textbf{60.82} \\
DivPrune & 25\% & 54.60 / 54.60 & 19.85 / 19.85 & 25.00 / 25.00 & 14.79 / 14.79 & 56.73 / \textbf{58.73} \\
DivPrune & 15\% & 44.41 / 44.41 & 9.88 / 9.88 & 15.00 / 15.00 & 9.11 / 9.11 & 55.30 / \textbf{58.06} \\
DivPrune & 10\% & 33.97 / 33.97 & 5.83 / 5.83 & 10.00 / 10.00 & 6.36 / 6.36 & 53.55 / \textbf{55.22} \\
DivPrune & 5\% & 17.01 / 17.01 & 2.91 / 2.91 & 5.00 / 5.00 & 3.68 / 3.68 & 49.62 / \textbf{52.38} \\
\midrule
DivPrune-M & 35\% & 64.64 / 64.64 & 29.84 / 29.84 & 35.00 / 35.00 & 20.70 / 20.70 & 59.06 / \textbf{61.15} \\
DivPrune-M & 25\% & 54.60 / 54.60 & 19.85 / 19.85 & 25.00 / 25.00 & 14.79 / 14.79 & 57.73 / \textbf{58.90} \\
DivPrune-M & 15\% & 44.41 / 44.41 & 9.88 / 9.88 & 15.00 / 15.00 & 9.11 / 9.11 & 55.39 / \textbf{58.23} \\
DivPrune-M & 10\% & 33.97 / 33.97 & 5.83 / 5.83 & 10.00 / 10.00 & 6.36 / 6.36 & 53.30 / \textbf{56.31} \\
DivPrune-M & 5\% & 17.01 / 17.01 & 2.91 / 2.91 & 5.00 / 5.00 & 3.68 / 3.68 & 50.29 / \textbf{51.96} \\
\midrule
OmniZip* & 35\% & 80.93 / 80.93 & 30.84 / 30.84 & 38.27 / 38.27 & 22.70 / 22.70 & 59.15 / \textbf{60.65} \\
OmniZip* & 25\% & 57.84 / 57.84 & 21.61 / 21.61 & 26.98 / 26.98 & 15.95 / 15.95 & 57.64 / \textbf{58.81} \\
OmniZip* & 15\% & 34.69 / 34.69 & 12.72 / 12.72 & 15.98 / 15.98 & 9.66 / 9.66 & 54.80 / \textbf{57.81} \\
OmniZip* & 10\% & 23.15 / 23.15 & 8.27 / 8.27 & 10.48 / 10.48 & 6.63 / 6.63 & 50.38 / \textbf{52.97} \\
OmniZip* & 5\% & 11.60 / 11.60 & 3.82 / 3.82 & 4.98 / 4.98 & 3.66 / 3.66 & 45.36 / \textbf{47.62} \\
\midrule
SEATS & 35\% & 65.00 / 65.00 & 30.00 / 30.00 & 35.19 / 35.19 & 19.86 / 19.86 & 61.32 / \textbf{62.24} \\
SEATS & 25\% & 55.00 / 55.00 & 20.00 / 20.00 & 25.19 / 25.19 & 14.13 / 14.13 & 59.90 / \textbf{60.48} \\
SEATS & 15\% & 45.00 / 45.00 & 10.00 / 10.00 & 15.19 / 15.19 & 8.73 / 8.73 & 56.64 / \textbf{58.65} \\
SEATS & 10\% & 35.00 / 35.00 & 6.00 / 6.00 & 10.30 / 10.30 & 6.09 / 6.09 & 56.31 / \textbf{56.89} \\
SEATS & 5\% & 9.44 / 9.44 & 3.15 / 3.15 & 4.08 / 4.08 & 3.10 / 3.10 & 48.71 / \textbf{50.63} \\
\midrule
\multicolumn{7}{l}{\textbf{AVUT}} \\*
Full tokens & 100\% & 100.00 & 100.00 & 100.00 & 79.94 & 64.65 \\*
\midrule
Uniform & 35\% & 34.97 / 34.97 & 35.01 / 35.01 & 35.00 / 35.00 & 24.50 / 24.50 & 60.44 / \textbf{62.86} \\
Uniform & 25\% & 25.00 / 25.00 & 25.00 / 25.00 & 25.00 / 25.00 & 17.31 / 17.31 & 56.81 / \textbf{59.05} \\
Uniform & 15\% & 15.00 / 15.00 & 15.00 / 15.00 & 15.00 / 15.00 & 10.49 / 10.49 & 52.94 / \textbf{55.25} \\
Uniform & 10\% & 10.00 / 10.00 & 10.00 / 10.00 & 10.00 / 10.00 & 7.20 / 7.20 & 50.35 / \textbf{51.50} \\
Uniform & 5\% & 5.05 / 5.05 & 4.99 / 4.99 & 5.00 / 5.00 & 4.01 / 4.01 & 46.37 / \textbf{47.92} \\
\midrule
DivPrune & 35\% & 59.60 / 59.60 & 28.86 / 28.86 & 35.00 / 35.00 & 24.50 / 24.50 & 62.28 / \textbf{63.21} \\
DivPrune & 25\% & 49.11 / 49.11 & 18.99 / 18.99 & 25.00 / 25.00 & 17.31 / 17.31 & 60.90 / \textbf{62.00} \\
DivPrune & 15\% & 38.14 / 38.14 & 9.23 / 9.23 & 15.00 / 15.00 & 10.49 / 10.49 & 58.19 / \textbf{60.03} \\
DivPrune & 10\% & 28.52 / 28.52 & 5.38 / 5.38 & 10.00 / 10.00 & 7.20 / 7.20 & 56.69 / \textbf{58.30} \\
DivPrune & 5\% & 14.26 / 14.26 & 2.69 / 2.69 & 5.00 / 5.00 & 4.01 / 4.01 & 52.88 / \textbf{54.27} \\
\midrule
DivPrune-M & 35\% & 59.60 / 59.60 & 28.86 / 28.86 & 35.00 / 35.00 & 24.50 / 24.50 & 62.46 / \textbf{63.44} \\
DivPrune-M & 25\% & 49.11 / 49.11 & 18.99 / 18.99 & 25.00 / 25.00 & 17.31 / 17.31 & 61.36 / \textbf{62.98} \\
DivPrune-M & 15\% & 38.14 / 38.14 & 9.23 / 9.23 & 15.00 / 15.00 & 10.49 / 10.49 & 59.52 / \textbf{60.09} \\
DivPrune-M & 10\% & 28.52 / 28.52 & 5.38 / 5.38 & 10.00 / 10.00 & 7.20 / 7.20 & 57.09 / \textbf{59.11} \\
DivPrune-M & 5\% & 14.26 / 14.26 & 2.69 / 2.69 & 5.00 / 5.00 & 4.01 / 4.01 & 53.92 / \textbf{55.25} \\
\midrule
OmniZip* & 35\% & 72.23 / 72.23 & 28.59 / 28.59 & 37.30 / 37.30 & 26.19 / 26.19 & 61.13 / \textbf{62.40} \\
OmniZip* & 25\% & 51.59 / 51.59 & 19.99 / 19.99 & 26.30 / 26.30 & 18.22 / 18.22 & 59.23 / \textbf{60.09} \\
OmniZip* & 15\% & 30.95 / 30.95 & 11.77 / 11.77 & 15.60 / 15.60 & 10.88 / 10.88 & 55.59 / \textbf{56.75} \\
OmniZip* & 10\% & 20.64 / 20.64 & 7.65 / 7.65 & 10.25 / 10.25 & 7.36 / 7.36 & 54.44 / \textbf{54.90} \\
OmniZip* & 5\% & 10.32 / 10.32 & 3.56 / 3.56 & 4.91 / 4.91 & 3.95 / 3.95 & 47.75 / \textbf{48.67} \\
\midrule
SEATS & 35\% & 65.00 / 65.00 & 30.00 / 30.00 & 36.99 / 36.99 & 25.23 / 25.23 & 64.94 / \textbf{65.97} \\
SEATS & 25\% & 55.00 / 55.00 & 20.00 / 20.00 & 26.99 / 26.99 & 18.05 / 18.05 & 62.69 / \textbf{63.55} \\
SEATS & 15\% & 45.00 / 45.00 & 10.00 / 10.00 & 16.99 / 16.99 & 11.41 / 11.41 & 59.52 / \textbf{60.84} \\
SEATS & 10\% & 35.00 / 35.00 & 6.00 / 6.00 & 11.79 / 11.79 & 7.89 / 7.89 & 57.90 / \textbf{58.77} \\
SEATS & 5\% & 9.44 / 9.44 & 3.15 / 3.15 & 4.40 / 4.40 & 3.55 / 3.55 & 49.65 / \textbf{51.61} \\
\midrule
\multicolumn{7}{l}{\textbf{OmniVideoBench}} \\*
Full tokens & 100\% & 100.00 & 100.00 & 100.00 & 248.99 & 35.50 \\*
\midrule
Uniform & 35\% & 34.99 / 34.99 & 35.00 / 35.00 & 35.00 / 35.00 & 65.15 / 65.15 & \textbf{33.80} / 33.70 \\
Uniform & 25\% & 25.01 / 25.01 & 25.00 / 25.00 & 25.00 / 25.00 & 44.32 / 44.32 & 31.40 / \textbf{32.60} \\
Uniform & 15\% & 14.99 / 14.99 & 15.00 / 15.00 & 15.00 / 15.00 & 25.48 / 25.48 & 29.90 / \textbf{31.50} \\
Uniform & 10\% & 10.00 / 10.00 & 10.00 / 10.00 & 10.00 / 10.00 & 16.81 / 16.81 & 29.50 / \textbf{30.80} \\
Uniform & 5\% & 5.01 / 5.01 & 5.00 / 5.00 & 5.00 / 5.00 & 8.63 / 8.63 & 30.10 / \textbf{31.00} \\
\midrule
DivPrune & 35\% & 57.94 / 57.94 & 27.12 / 27.12 & 35.00 / 35.00 & 65.15 / 65.15 & 34.90 / \textbf{35.50} \\
DivPrune & 25\% & 47.03 / 47.03 & 17.43 / 17.43 & 25.00 / 25.00 & 44.32 / 44.32 & 34.30 / \textbf{35.00} \\
DivPrune & 15\% & 35.19 / 35.19 & 8.06 / 8.06 & 15.00 / 15.00 & 25.48 / 25.48 & 32.30 / \textbf{34.70} \\
DivPrune & 10\% & 25.76 / 25.76 & 4.58 / 4.58 & 10.00 / 10.00 & 16.81 / 16.81 & 33.60 / \textbf{35.30} \\
DivPrune & 5\% & 12.88 / 12.88 & 2.29 / 2.29 & 5.00 / 5.00 & 8.63 / 8.63 & 31.50 / \textbf{34.20} \\
\midrule
DivPrune-M & 35\% & 57.94 / 57.94 & 27.12 / 27.12 & 35.00 / 35.00 & 65.15 / 65.15 & 34.70 / \textbf{35.70} \\
DivPrune-M & 25\% & 47.03 / 47.03 & 17.43 / 17.43 & 25.00 / 25.00 & 44.32 / 44.32 & 34.20 / \textbf{35.40} \\
DivPrune-M & 15\% & 35.19 / 35.19 & 8.06 / 8.06 & 15.00 / 15.00 & 25.48 / 25.48 & 33.50 / \textbf{34.70} \\
DivPrune-M & 10\% & 25.76 / 25.76 & 4.58 / 4.58 & 10.00 / 10.00 & 16.81 / 16.81 & 33.90 / \textbf{34.60} \\
DivPrune-M & 5\% & 12.88 / 12.88 & 2.29 / 2.29 & 5.00 / 5.00 & 8.63 / 8.63 & 32.50 / \textbf{32.70} \\
\midrule
OmniZip* & 35\% & 68.74 / 68.74 & 25.69 / 25.69 & 36.70 / 36.70 & 68.82 / 68.82 & \textbf{35.00} / 34.70 \\
OmniZip* & 25\% & 49.10 / 49.10 & 18.05 / 18.05 & 25.99 / 25.99 & 46.27 / 46.27 & 31.80 / \textbf{34.00} \\
OmniZip* & 15\% & 29.46 / 29.46 & 10.58 / 10.58 & 15.41 / 15.41 & 26.20 / 26.20 & 33.00 / \textbf{33.30} \\
OmniZip* & 10\% & 19.64 / 19.64 & 6.95 / 6.95 & 10.19 / 10.19 & 17.13 / 17.13 & 31.70 / \textbf{32.80} \\
OmniZip* & 5\% & 9.82 / 9.82 & 3.15 / 3.15 & 4.85 / 4.85 & 8.40 / 8.40 & 30.50 / \textbf{31.60} \\
\midrule
SEATS & 35\% & 65.00 / 65.00 & 30.00 / 30.00 & 38.95 / 38.95 & 74.36 / 74.36 & \textbf{35.70} / 35.50 \\
SEATS & 25\% & 55.00 / 55.00 & 20.00 / 20.00 & 28.95 / 28.95 & 51.75 / 51.75 & \textbf{35.40} / 35.00 \\
SEATS & 15\% & 45.00 / 45.00 & 10.00 / 10.00 & 18.95 / 18.95 & 32.13 / 32.13 & \textbf{36.00} / 35.70 \\
SEATS & 10\% & 35.00 / 35.00 & 6.00 / 6.00 & 13.42 / 13.42 & 21.48 / 21.48 & 34.30 / \textbf{36.20} \\
SEATS & 5\% & 9.44 / 9.44 & 3.15 / 3.15 & 4.75 / 4.75 & 8.08 / 8.08 & 31.20 / \textbf{31.90} \\
\midrule
\multicolumn{7}{l}{\textbf{Video-MME}} \\*
Full tokens & 100\% & 100.00 & 100.00 & 100.00 & 839.14 & 64.41 \\*
\midrule
Uniform & 35\% & 35.00 / 35.00 & 35.00 / 35.00 & 35.00 / 35.00 & 176.99 / 176.99 & 63.44 / \textbf{64.81} \\
Uniform & 25\% & 25.00 / 25.00 & 25.00 / 25.00 & 25.00 / 25.00 & 113.80 / 113.80 & 61.63 / \textbf{62.70} \\
Uniform & 15\% & 15.00 / 15.00 & 15.00 / 15.00 & 15.00 / 15.00 & 60.92 / 60.92 & 60.33 / \textbf{61.67} \\
Uniform & 10\% & 10.00 / 10.00 & 10.00 / 10.00 & 10.00 / 10.00 & 38.35 / 38.35 & 58.15 / \textbf{60.22} \\
Uniform & 5\% & 5.01 / 5.01 & 5.00 / 5.00 & 5.00 / 5.00 & 18.36 / 18.36 & 54.89 / \textbf{57.37} \\
\midrule
DivPrune & 35\% & 66.14 / 66.14 & 30.56 / 30.56 & 35.00 / 35.00 & 176.99 / 176.99 & 63.15 / \textbf{64.44} \\
DivPrune & 25\% & 56.37 / 56.37 & 20.53 / 20.53 & 25.00 / 25.00 & 113.80 / 113.80 & 62.04 / \textbf{63.15} \\
DivPrune & 15\% & 46.90 / 46.90 & 10.45 / 10.45 & 15.00 / 15.00 & 60.92 / 60.92 & 61.15 / \textbf{62.89} \\
DivPrune & 10\% & 36.32 / 36.32 & 6.25 / 6.25 & 10.00 / 10.00 & 38.35 / 38.35 & 59.56 / \textbf{61.04} \\
DivPrune & 5\% & 18.16 / 18.16 & 3.12 / 3.12 & 5.00 / 5.00 & 18.36 / 18.36 & 56.48 / \textbf{59.52} \\
\midrule
DivPrune-M & 35\% & 66.14 / 66.14 & 30.56 / 30.56 & 35.00 / 35.00 & 176.99 / 176.99 & 63.30 / \textbf{64.33} \\
DivPrune-M & 25\% & 56.37 / 56.37 & 20.53 / 20.53 & 25.00 / 25.00 & 113.80 / 113.80 & 62.41 / \textbf{64.00} \\
DivPrune-M & 15\% & 46.90 / 46.90 & 10.45 / 10.45 & 15.00 / 15.00 & 60.92 / 60.92 & 61.11 / \textbf{62.63} \\
DivPrune-M & 10\% & 36.32 / 36.32 & 6.25 / 6.25 & 10.00 / 10.00 & 38.35 / 38.35 & 59.96 / \textbf{61.74} \\
DivPrune-M & 5\% & 18.16 / 18.16 & 3.12 / 3.12 & 5.00 / 5.00 & 18.36 / 18.36 & 57.07 / \textbf{59.48} \\
\midrule
OmniZip* & 35\% & 83.93 / 83.93 & 29.33 / 29.33 & 36.15 / 36.15 & 184.28 / 184.28 & 63.07 / \textbf{64.26} \\
OmniZip* & 25\% & 59.94 / 59.94 & 20.89 / 20.89 & 25.76 / 25.76 & 118.03 / 118.03 & 62.89 / \textbf{63.74} \\
OmniZip* & 15\% & 35.97 / 35.97 & 12.28 / 12.28 & 15.23 / 15.23 & 61.96 / 61.96 & 61.04 / \textbf{61.81} \\
OmniZip* & 10\% & 23.98 / 23.98 & 7.97 / 7.97 & 9.96 / 9.96 & 38.16 / 38.16 & 59.04 / \textbf{60.56} \\
OmniZip* & 5\% & 11.99 / 11.99 & 3.66 / 3.66 & 4.70 / 4.70 & 17.23 / 17.23 & 55.04 / \textbf{56.93} \\
\midrule
SEATS & 35\% & 65.00 / 65.00 & 30.00 / 30.00 & 34.37 / 34.37 & 183.19 / 183.19 & 65.44 / \textbf{66.19} \\
SEATS & 25\% & 55.00 / 55.00 & 20.00 / 20.00 & 24.37 / 24.37 & 113.84 / 113.84 & 64.85 / \textbf{65.96} \\
SEATS & 15\% & 45.00 / 45.00 & 10.00 / 10.00 & 14.37 / 14.37 & 57.81 / 57.81 & 63.26 / \textbf{64.22} \\
SEATS & 10\% & 35.00 / 35.00 & 6.00 / 6.00 & 9.62 / 9.62 & 35.30 / 35.30 & 61.48 / \textbf{62.48} \\
SEATS & 5\% & 9.44 / 9.44 & 3.15 / 3.15 & 3.93 / 3.93 & 14.19 / 14.19 & 56.44 / \textbf{58.30} \\
 
\end{longtable}
\endgroup

\subsection{Training Data Scale and Retention}
\label{app:training_condition_details}

Table~\ref{tab:data_scale_details} gives every WorldSense deployment result
for the data-scale study. All rows use two epochs, so the 10\%, 30\%, 50\%,
and 100\% subsets contain 570, 1,710, 2,850, and 5,700 instances and produce
36, 108, 180, and 358 optimizer updates, respectively. The nested subsets
therefore change both the number of distinct examples and optimization steps.

\begingroup
\appendixlongtablestyle
\setlength{\tabcolsep}{5pt}
\begin{longtable}{llrrrrrr}
\caption{WorldSense accuracy (\%) for the DivPrune data-scale by
training-retention grid.}
\label{tab:data_scale_details}\\
\toprule
Training data & Train & Eval35 & Eval25 & Eval15 & Eval10 & Eval5 & Avg. \\
\midrule
\endfirsthead
\multicolumn{8}{c}{\tablename\ \thetable\ continued}\\
\toprule
Training data & Train & Eval35 & Eval25 & Eval15 & Eval10 & Eval5 & Avg. \\
\midrule
\endhead
\midrule
\multicolumn{8}{r}{Continued on next page}\\
\endfoot
\bottomrule
\endlastfoot
100\% (5700) & 35\% & 45.87 & 45.52 & 44.07 & 42.34 & 39.06 & 43.37 \\
 & 25\% & 46.15 & 45.55 & 44.01 & 42.81 & 39.34 & 43.58 \\
 & 15\% & 46.31 & 45.78 & 44.26 & 42.91 & 39.66 & 43.78 \\
 & 10\% & 46.15 & 46.06 & 44.42 & 42.97 & 40.04 & 43.93 \\
 & 5\% & 46.37 & 46.15 & 44.45 & 43.35 & 40.45 & 44.16 \\
\midrule
50\% (2850) & 35\% & 45.46 & 45.08 & 43.66 & 42.50 & 38.93 & 43.13 \\
 & 25\% & 45.71 & 44.89 & 44.10 & 42.62 & 39.09 & 43.28 \\
 & 15\% & 45.74 & 44.92 & 43.92 & 42.75 & 39.09 & 43.28 \\
 & 10\% & 45.71 & 45.15 & 43.85 & 42.84 & 39.28 & 43.37 \\
 & 5\% & 45.93 & 45.78 & 44.07 & 43.16 & 40.10 & 43.81 \\
\midrule
30\% (1710) & 35\% & 45.65 & 45.11 & 43.41 & 42.84 & 38.97 & 43.20 \\
 & 25\% & 45.71 & 45.21 & 43.66 & 42.40 & 39.09 & 43.22 \\
 & 15\% & 45.65 & 45.18 & 43.60 & 42.59 & 38.93 & 43.19 \\
 & 10\% & 45.71 & 45.30 & 43.60 & 42.34 & 39.19 & 43.23 \\
 & 5\% & 45.71 & 45.18 & 43.82 & 42.62 & 39.50 & 43.37 \\
\midrule
10\% (570) & 35\% & 45.27 & 45.21 & 43.44 & 42.28 & 39.00 & 43.04 \\
 & 25\% & 45.27 & 44.96 & 43.10 & 42.15 & 39.19 & 42.93 \\
 & 15\% & 45.33 & 45.11 & 43.41 & 42.15 & 39.09 & 43.02 \\
 & 10\% & 45.49 & 45.02 & 43.28 & 42.06 & 39.31 & 43.03 \\
 & 5\% & 45.43 & 45.15 & 43.63 & 42.40 & 38.97 & 43.11 \\
 
\end{longtable}
\endgroup

Table~\ref{tab:train_retention_details} reports all per-budget accuracies underlying
Figure~\ref{fig:train_rtt_and_datacount}(b). Each checkpoint is evaluated on
three benchmarks at all five deployment budgets. Comparisons across
compressors are not equal-compute comparisons because their retention
semantics differ.

\begingroup
\appendixlongtablestyle
\setlength{\tabcolsep}{5pt}
\begin{longtable}{llrrrrrrr}
\caption{Accuracy (\%) across training and deployment retention.}
\label{tab:train_retention_details}\\
\toprule
Compressor & Train & Benchmark & Eval35 & Eval25 & Eval15 & Eval10 & Eval5 & Avg. \\
\midrule
\endfirsthead
\multicolumn{9}{c}{\tablename\ \thetable\ continued}\\
\toprule
Compressor & Train & Benchmark & Eval35 & Eval25 & Eval15 & Eval10 & Eval5 & Avg. \\
\midrule
\endhead
\midrule
\multicolumn{9}{r}{Continued on next page}\\
\endfoot
\bottomrule
\endlastfoot
DivPrune & 35\% & WorldSense & 45.87 & 45.52 & 44.07 & 42.34 & 39.06 & 43.37 \\
 &  & DailyOmni & 60.65 & 57.81 & 56.31 & 53.97 & 50.79 & 55.91 \\
 &  & AVUT & 62.57 & 61.48 & 59.46 & 57.79 & 53.34 & 58.93 \\
\addlinespace[1pt]
 & 25\% & WorldSense & 46.15 & 45.55 & 44.01 & 42.81 & 39.34 & 43.58 \\
 &  & DailyOmni & 60.90 & 58.15 & 56.64 & 54.39 & 50.96 & 56.21 \\
 &  & AVUT & 62.98 & 61.42 & 59.34 & 58.02 & 53.46 & 59.04 \\
\addlinespace[1pt]
 & 15\% & WorldSense & 46.31 & 45.78 & 44.26 & 42.91 & 39.66 & 43.78 \\
 &  & DailyOmni & 60.74 & 58.56 & 57.48 & 54.22 & 51.38 & 56.47 \\
 &  & AVUT & 62.98 & 61.65 & 59.92 & 58.19 & 53.40 & 59.23 \\
\addlinespace[1pt]
 & 10\% & WorldSense & 46.15 & 46.06 & 44.42 & 42.97 & 40.04 & 43.93 \\
 &  & DailyOmni & 61.15 & 58.48 & 57.73 & 54.89 & 51.71 & 56.79 \\
 &  & AVUT & 63.44 & 61.88 & 59.86 & 58.19 & 53.63 & 59.40 \\
\addlinespace[1pt]
 & 5\% & WorldSense & 46.37 & 46.15 & 44.45 & 43.35 & 40.45 & 44.16 \\
 &  & DailyOmni & 60.82 & 58.73 & 58.06 & 55.22 & 52.38 & 57.04 \\
 &  & AVUT & 63.21 & 62.00 & 60.03 & 58.30 & 54.27 & 59.56 \\
\addlinespace[1pt]
 & 3\% & WorldSense & 46.00 & 46.34 & 44.64 & 43.41 & 40.32 & 44.14 \\
 &  & DailyOmni & 60.23 & 58.90 & 58.15 & 55.47 & 51.96 & 56.94 \\
 &  & AVUT & 63.26 & 62.00 & 59.75 & 58.19 & 54.33 & 59.50 \\
\addlinespace[1pt]
 & 1\% & WorldSense & 45.90 & 46.06 & 44.20 & 42.97 & 39.94 & 43.81 \\
 &  & DailyOmni & 61.15 & 58.15 & 57.64 & 55.05 & 52.30 & 56.86 \\
 &  & AVUT & 63.44 & 61.48 & 60.55 & 58.07 & 54.44 & 59.60 \\
\midrule
OmniZip* & 35\% & WorldSense & 44.80 & 43.76 & 40.70 & 39.03 & 35.28 & 40.71 \\
 &  & DailyOmni & 59.73 & 58.48 & 56.39 & 52.38 & 45.95 & 54.59 \\
 &  & AVUT & 61.53 & 59.17 & 56.17 & 54.44 & 48.04 & 55.87 \\
\addlinespace[1pt]
 & 25\% & WorldSense & 45.11 & 44.01 & 40.92 & 39.34 & 35.40 & 40.96 \\
 &  & DailyOmni & 60.07 & 58.40 & 57.14 & 52.13 & 46.62 & 54.87 \\
 &  & AVUT & 61.71 & 59.40 & 56.11 & 54.61 & 48.04 & 55.97 \\
\addlinespace[1pt]
 & 15\% & WorldSense & 45.37 & 44.20 & 41.11 & 39.79 & 35.53 & 41.20 \\
 &  & DailyOmni & 60.32 & 59.06 & 57.31 & 52.38 & 47.12 & 55.24 \\
 &  & AVUT & 62.05 & 59.86 & 56.29 & 54.67 & 48.21 & 56.22 \\
\addlinespace[1pt]
 & 10\% & WorldSense & 45.65 & 44.58 & 41.36 & 39.94 & 36.13 & 41.53 \\
 &  & DailyOmni & 60.65 & 58.65 & 58.06 & 52.21 & 47.37 & 55.39 \\
 &  & AVUT & 62.17 & 59.75 & 56.34 & 55.19 & 48.96 & 56.48 \\
\addlinespace[1pt]
 & 5\% & WorldSense & 45.71 & 44.58 & 41.24 & 40.64 & 36.48 & 41.73 \\
 &  & DailyOmni & 60.65 & 58.81 & 57.81 & 52.97 & 47.62 & 55.57 \\
 &  & AVUT & 62.40 & 60.09 & 56.75 & 54.90 & 48.67 & 56.56 \\
\addlinespace[1pt]
 & 3\% & WorldSense & 45.59 & 44.33 & 41.14 & 40.16 & 36.95 & 41.63 \\
 &  & DailyOmni & 60.82 & 58.81 & 57.39 & 53.22 & 47.79 & 55.61 \\
 &  & AVUT & 62.40 & 59.98 & 57.27 & 54.67 & 48.85 & 56.63 \\
\addlinespace[1pt]
 & 1\% & WorldSense & 45.33 & 43.98 & 40.98 & 39.50 & 36.60 & 41.28 \\
 &  & DailyOmni & 60.57 & 58.73 & 56.47 & 51.80 & 47.54 & 55.02 \\
 &  & AVUT & 62.05 & 60.38 & 56.92 & 54.73 & 48.62 & 56.54 \\
\midrule
SEATS & 35\% & WorldSense & 46.78 & 46.34 & 44.80 & 43.57 & 38.34 & 43.97 \\
 &  & DailyOmni & 61.57 & 60.57 & 56.98 & 56.47 & 48.87 & 56.89 \\
 &  & AVUT & 65.05 & 62.80 & 59.80 & 58.07 & 50.29 & 59.20 \\
\addlinespace[1pt]
 & 25\% & WorldSense & 46.97 & 46.37 & 44.96 & 43.69 & 38.18 & 44.04 \\
 &  & DailyOmni & 61.90 & 60.65 & 57.06 & 56.64 & 49.54 & 57.16 \\
 &  & AVUT & 65.11 & 62.57 & 59.57 & 58.13 & 50.75 & 59.23 \\
\addlinespace[1pt]
 & 15\% & WorldSense & 47.04 & 46.53 & 45.18 & 43.63 & 37.93 & 44.06 \\
 &  & DailyOmni & 61.99 & 60.23 & 57.14 & 56.06 & 49.79 & 57.04 \\
 &  & AVUT & 64.99 & 62.92 & 59.98 & 58.19 & 50.69 & 59.35 \\
\addlinespace[1pt]
 & 10\% & WorldSense & 47.10 & 46.60 & 45.05 & 43.88 & 38.46 & 44.22 \\
 &  & DailyOmni & 61.82 & 60.32 & 57.39 & 56.39 & 50.13 & 57.21 \\
 &  & AVUT & 65.17 & 62.80 & 59.98 & 58.25 & 50.92 & 59.42 \\
\addlinespace[1pt]
 & 5\% & WorldSense & 47.51 & 47.32 & 45.62 & 44.10 & 39.31 & 44.77 \\
 &  & DailyOmni & 62.24 & 60.48 & 58.65 & 56.89 & 50.63 & 57.78 \\
 &  & AVUT & 65.97 & 63.55 & 60.84 & 58.77 & 51.61 & 60.15 \\
\addlinespace[1pt]
 & 3\% & WorldSense & 47.60 & 47.07 & 45.84 & 44.39 & 39.75 & 44.93 \\
 &  & DailyOmni & 62.16 & 60.65 & 58.48 & 57.14 & 50.54 & 57.79 \\
 &  & AVUT & 65.97 & 63.32 & 60.90 & 58.59 & 51.15 & 59.99 \\
\addlinespace[1pt]
 & 1\% & WorldSense & 47.13 & 46.85 & 45.08 & 43.92 & 39.47 & 44.49 \\
 &  & DailyOmni & 61.90 & 60.32 & 58.06 & 57.31 & 50.21 & 57.56 \\
 &  & AVUT & 65.69 & 63.38 & 60.55 & 58.59 & 51.79 & 60.00 \\
 
\end{longtable}
\endgroup

\subsection{Distillation Objective}
\label{app:objective_details}

Table~\ref{tab:objective_budget_details} expands the distillation-objective
study in Table~\ref{tab:distillation_design} to every deployment budget.
The three settings share the same training configuration; only the
vocabulary-level divergence changes. Forward KL is
\(\mathrm{KL}(p_T\|p_S)\), while reverse KL is
\(\mathrm{KL}(p_S\|p_T)\).
All three objectives use the same pointwise clipping threshold of 0.05.

\begingroup
\appendixlongtablestyle
\setlength{\tabcolsep}{5pt}
\begin{longtable}{llrrrrrr}
\caption{\textbf{Per-budget distillation-objective results.}
All runs use DivPrune at 5\% training retention and are evaluated with
DivPrune at the five listed budgets. Accuracies (\%) are reported for
WorldSense, DailyOmni, and AVUT. The bold CAFD entry uses JSD and
the same seed-42 checkpoint as Table~\ref{tab:main_results}.}
\label{tab:objective_budget_details}\\
\toprule
Variant & Benchmark & 35\% & 25\% & 15\% & 10\% & 5\% & Avg. \\
\midrule
\endfirsthead
\multicolumn{8}{c}{\tablename\ \thetable\ continued}\\
\toprule
Variant & Benchmark & 35\% & 25\% & 15\% & 10\% & 5\% & Avg. \\
\midrule
\endhead
\midrule
\multicolumn{8}{r}{Continued on next page}\\
\endfoot
\bottomrule
\endlastfoot
\textbf{CAFD} & WorldSense & 46.37 & 46.15 & 44.45 & 43.35 & 40.45 & 44.16 \\*
(JSD) & DailyOmni & 60.82 & 58.73 & 58.06 & 55.22 & 52.38 & 57.04 \\*
 & AVUT & 63.21 & 62.00 & 60.03 & 58.30 & 54.27 & 59.56 \\
\addlinespace
Forward KL & WorldSense & 44.80 & 44.39 & 42.59 & 41.58 & 38.65 & 42.40 \\*
 & DailyOmni & 58.90 & 57.06 & 55.22 & 53.63 & 49.54 & 54.87 \\*
 & AVUT & 62.00 & 60.38 & 58.94 & 56.92 & 53.23 & 58.29 \\
\addlinespace
Reverse KL & WorldSense & 45.71 & 45.37 & 43.69 & 42.43 & 39.06 & 43.25 \\*
 & DailyOmni & 59.57 & 57.48 & 55.64 & 53.05 & 51.13 & 55.37 \\*
 & AVUT & 62.46 & 61.76 & 59.46 & 57.50 & 53.69 & 58.97 \\
 
\end{longtable}
\endgroup

\subsection{Task Conditioning and Supervision}
\label{app:control_details}

Table~\ref{tab:control_budget_details} expands the task-conditioning and
supervision studies in Tables~\ref{tab:supervision_controls}
and~\ref{tab:app_supervision_controls} to every deployment budget.
These results show that the main configuration is not uniformly the most
accurate. CAFD (question-only) slightly outperforms the main configuration on
average, suggesting that including candidate options during adaptation does
not necessarily improve accuracy.

\begingroup
\appendixlongtablestyle
\setlength{\tabcolsep}{5pt}
\begin{longtable}{llrrrrrr}
\caption{\textbf{Per-budget task-conditioning and supervision results.}
All adapted variants use DivPrune at 5\% training retention and are
evaluated with DivPrune at the five listed budgets. Accuracies (\%) are
reported for WorldSense, DailyOmni, and AVUT. The fixed-generic and
question-only rows are reference-free CAFD variants. The bold
CAFD entry denotes the main configuration used in
Table~\ref{tab:main_results}, with its task input in parentheses.}
\label{tab:control_budget_details}\\
\toprule
Variant & Benchmark & 35\% & 25\% & 15\% & 10\% & 5\% & Avg. \\
\midrule
\endfirsthead
\multicolumn{8}{c}{\tablename\ \thetable\ continued}\\
\toprule
Variant & Benchmark & 35\% & 25\% & 15\% & 10\% & 5\% & Avg. \\
\midrule
\endhead
\midrule
\multicolumn{8}{r}{Continued on next page}\\
\endfoot
\bottomrule
\endlastfoot
Unadapted & WorldSense & 45.49 & 45.02 & 43.25 & 42.06 & 39.12 & 42.99 \\*
 & DailyOmni & 58.48 & 56.73 & 55.30 & 53.55 & 49.62 & 54.74 \\*
 & AVUT & 62.28 & 60.90 & 58.19 & 56.69 & 52.88 & 58.19 \\
\addlinespace
\midrule
CAFD (fixed-generic) & WorldSense & 46.09 & 46.31 & 44.80 & 43.38 & 40.45 & 44.21 \\*
 & DailyOmni & 61.32 & 59.06 & 56.98 & 55.47 & 52.30 & 57.03 \\*
 & AVUT & 63.32 & 61.94 & 59.52 & 57.55 & 53.63 & 59.19 \\
\addlinespace
CAFD (question-only) & WorldSense & 46.28 & 46.34 & 44.80 & 43.44 & 40.07 & 44.19 \\*
 & DailyOmni & 60.65 & 59.06 & 57.98 & 55.72 & 52.88 & 57.26 \\*
 & AVUT & 63.84 & 61.59 & 59.75 & 58.54 & 54.61 & 59.67 \\
\addlinespace
\textbf{CAFD} & WorldSense & 46.37 & 46.15 & 44.45 & 43.35 & 40.45 & 44.16 \\*
(Question + options) & DailyOmni & 60.82 & 58.73 & 58.06 & 55.22 & 52.38 & 57.04 \\*
 & AVUT & 63.21 & 62.00 & 60.03 & 58.30 & 54.27 & 59.56 \\
\addlinespace
\midrule
CAFD + reference & WorldSense & 46.41 & 45.93 & 44.67 & 43.22 & 40.61 & 44.17 \\*
 & DailyOmni & 60.40 & 58.90 & 57.89 & 55.05 & 51.04 & 56.66 \\*
 & AVUT & 63.09 & 61.94 & 59.69 & 58.13 & 54.61 & 59.49 \\
\addlinespace
Compressed-input SFT & WorldSense & 46.53 & 46.37 & 44.67 & 43.35 & 40.64 & 44.31 \\*
 & DailyOmni & 61.57 & 59.90 & 58.48 & 56.89 & 52.05 & 57.78 \\*
 & AVUT & 64.53 & 62.34 & 60.27 & 58.25 & 54.09 & 59.90 \\
 
\end{longtable}
\endgroup

\subsection{Training Dynamics}
\label{app:loss_dynamics}

\begin{figure}[h]
    \centering
    \includegraphics[width=\linewidth]{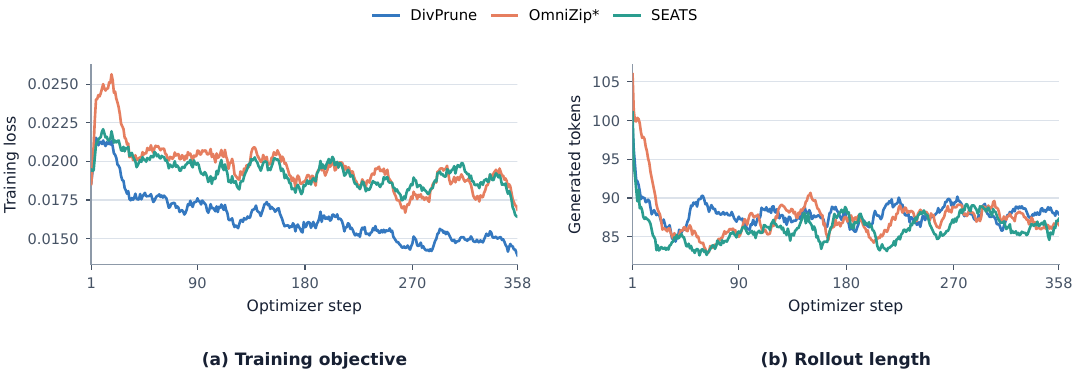}
    \caption{\textbf{Training dynamics at 5\% training retention.}
    (a) Training loss (clipped JSD). (b) Student rollout length.
    Curves show 15-step trailing averages for DivPrune, OmniZip*, and SEATS.}
    \label{fig:loss_dynamics}
\end{figure}

Figure~\ref{fig:loss_dynamics} visualizes the logged objective and
rollout length for the available final DivPrune, OmniZip*, and SEATS
checkpoints. The training loss caps each vocabulary-level JSD contribution
at 0.05 before summing over the vocabulary and averaging over generated
response positions. Contributions above the threshold still count as 0.05
in the loss, but their local gradient through clipping is zero.

\begingroup
\section{Behavioral Analysis}
\label{app:behavior_details}

We examine changes in individual predictions and confidence, and break down
results by task type.

\subsection{Repair and Regression}
\label{app:repair_regression_details}
\label{sec:repair_regression}

We examine paired DivPrune predictions using the Base full-token result
\(F_0\), the Unadapted compressed result \(C_0\), and the adapted compressed
result \(C_1\). A repair changes an incorrect compressed prediction to a
correct one; a regression changes a correct prediction to an incorrect one.

\begin{figure}[htbp]
    \centering
    \includegraphics[width=\linewidth]{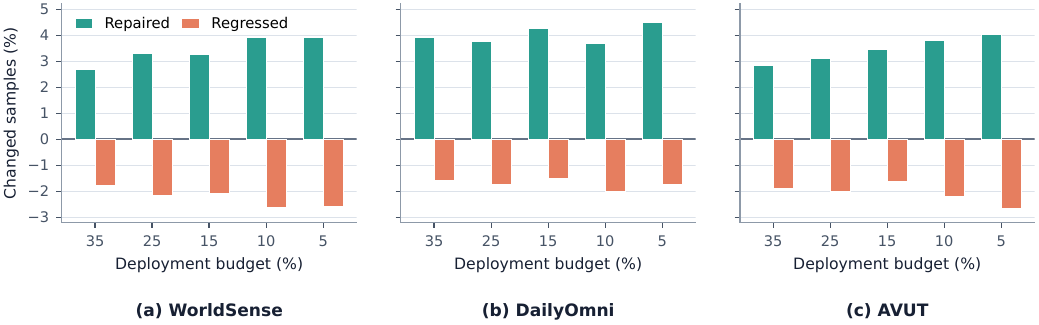}
    \caption{\textbf{Paired repair and regression under compression.}
    Bars show the percentages of matched examples changing from incorrect to
    correct (positive) or correct to incorrect (negative) after adaptation.
    Percentages use all evaluated examples as the denominator.
    Each panel reports all five DivPrune deployment budgets.}
    \label{fig:repair_regression}
\end{figure}

Table~\ref{tab:repair_details} compares each example's predictions before
and after adaptation under the same compression setting.
Wrong$\to$Correct counts corrected predictions, while Correct$\to$Wrong
counts newly incorrect predictions. Each cell reports the count and its
percentage of all evaluated examples. The last column reports the net
increase in correct predictions and the corresponding accuracy gain.
Across all three benchmarks and five deployment budgets, corrected predictions
consistently outnumber newly incorrect predictions, yielding positive accuracy
gains in every condition. DailyOmni shows the largest net accuracy gains across
all five budgets. These results show that the aggregate improvements reflect
a consistent excess of corrections over regressions.

\begin{table}[htbp]
    \centering
    \appendixtablestyle
    \caption{Matched prediction transitions for DivPrune. The model is
    trained with DivPrune at 5\% nominal retention and evaluated with
    DivPrune at the listed budgets. Transition cells report count / percentage
    of all evaluated examples; $\Delta$ Acc. reports net correct count /
    accuracy gain in percentage points (pp).}
    \label{tab:repair_details}
    \begin{autowidthtabular}{lrrrr}
        \toprule
        Benchmark & Budget & \shortstack{Wrong$\to$Correct\\(count/\%)} & \shortstack{Correct$\to$Wrong\\(count/\%)} & \shortstack{$\Delta$ Acc.\\(count/pp)} \\
        \midrule
        WorldSense & 35\% & 85 / 2.68 & 57 / 1.80 & 28 / 0.88 \\
WorldSense & 25\% & 105 / 3.31 & 69 / 2.18 & 36 / 1.13 \\
WorldSense & 15\% & 104 / 3.28 & 66 / 2.08 & 38 / 1.20 \\
WorldSense & 10\% & 124 / 3.91 & 83 / 2.62 & 41 / 1.29 \\
WorldSense & 5\% & 124 / 3.91 & 82 / 2.59 & 42 / 1.32 \\
DailyOmni & 35\% & 47 / 3.93 & 19 / 1.59 & 28 / 2.34 \\
DailyOmni & 25\% & 45 / 3.76 & 21 / 1.75 & 24 / 2.01 \\
DailyOmni & 15\% & 51 / 4.26 & 18 / 1.50 & 33 / 2.76 \\
DailyOmni & 10\% & 44 / 3.68 & 24 / 2.01 & 20 / 1.67 \\
DailyOmni & 5\% & 54 / 4.51 & 21 / 1.75 & 33 / 2.76 \\
AVUT & 35\% & 49 / 2.83 & 33 / 1.90 & 16 / 0.92 \\
AVUT & 25\% & 54 / 3.11 & 35 / 2.02 & 19 / 1.10 \\
AVUT & 15\% & 60 / 3.46 & 28 / 1.61 & 32 / 1.85 \\
AVUT & 10\% & 66 / 3.81 & 38 / 2.19 & 28 / 1.61 \\
AVUT & 5\% & 70 / 4.04 & 46 / 2.65 & 24 / 1.38 \\
\bottomrule
    \end{autowidthtabular}
\end{table}

\subsection{Confidence and Recovery Diagnostics}
\label{app:confidence_diagnostics}

\paragraph{Correct-option scores.}
The first three metric columns in Table~\ref{tab:continuous_diagnostics}
measure whether adaptation shifts the compressed model's scores toward the
correct answer. At the first response position, we normalize the logits over
the question's candidate option labels to obtain \(p(y)\), the probability
of the correct option \(y\). We also measure its natural log-probability
\(\log p(y)\) and logit margin:
\[
    \mathrm{margin}=z_y-\max_{k\ne y}z_k,
\]
where \(z_k\) is the logit of candidate option \(k\). The margin measures
how far the correct option leads or trails the highest-scoring incorrect
option. Each \(\Delta\) column averages the per-example change from
Unadapted compressed input \(C_0\) to adapted compressed input \(C_1\).
All three average changes are positive at every budget on all three
benchmarks: adaptation improves both correct-option probabilities and
relative scores, including changes that do not alter the final answer.

\paragraph{Error recovery and accuracy-gap recovery.}
The final two columns compare these predictions with the unadapted
full-token reference \(F_0\). Compression-error recovery (Comp.-error rec.)
considers only questions that \(F_0\) answers correctly but \(C_0\) answers
incorrectly, and reports the percentage of those questions answered correctly
by \(C_1\). RCG instead measures the fraction of the overall accuracy gap
between full and compressed inputs recovered by adaptation:
\[
    \mathrm{RCG}=100\times
    \frac{\mathrm{Acc}(C_1)-\mathrm{Acc}(C_0)}
         {\mathrm{Acc}(F_0)-\mathrm{Acc}(C_0)}.
\]
We report RCG when the denominator is positive. The two percentages answer
different questions: compression-error recovery tracks repairs within a
specific subset, whereas RCG uses the net accuracy gain over all examples,
including both corrections and regressions.

\begin{table}[htbp]
    \centering
    \appendixtablestyle
    \caption{Continuous paired diagnostics and recovery measures. The
    model is trained with DivPrune at 5\% nominal retention; compressed
    evaluations use DivPrune at the listed budgets. The final two columns
    are percentages.}
    \label{tab:continuous_diagnostics}
    \begin{autowidthtabular}{lrrrrrr}
        \toprule
        Benchmark & Budget & \(\Delta p(y)\) & \(\Delta\log p(y)\) & \(\Delta\)margin & Comp.-error rec. & RCG \\
        \midrule
        WorldSense & 35\% & 0.0060 & 0.0335 & 0.0555 & 27.91 & 65.12 \\
WorldSense & 25\% & 0.0071 & 0.0377 & 0.0636 & 28.77 & 62.07 \\
WorldSense & 15\% & 0.0091 & 0.0437 & 0.0735 & 23.94 & 33.33 \\
WorldSense & 10\% & 0.0103 & 0.0485 & 0.0831 & 22.32 & 26.97 \\
WorldSense & 5\% & 0.0115 & 0.0475 & 0.0967 & 18.44 & 17.14 \\
DailyOmni & 35\% & 0.0126 & 0.0749 & 0.1060 & 32.94 & 52.83 \\
DailyOmni & 25\% & 0.0149 & 0.0816 & 0.1337 & 30.28 & 32.43 \\
DailyOmni & 15\% & 0.0174 & 0.0847 & 0.1495 & 29.46 & 36.26 \\
DailyOmni & 10\% & 0.0169 & 0.0883 & 0.1513 & 21.88 & 17.86 \\
DailyOmni & 5\% & 0.0172 & 0.0827 & 0.1659 & 20.98 & 20.75 \\
AVUT & 35\% & 0.0077 & 0.0392 & 0.0637 & 27.62 & 39.02 \\
AVUT & 25\% & 0.0082 & 0.0389 & 0.0684 & 27.27 & 29.23 \\
AVUT & 15\% & 0.0080 & 0.0393 & 0.0676 & 21.93 & 28.57 \\
AVUT & 10\% & 0.0076 & 0.0372 & 0.0716 & 22.12 & 20.29 \\
AVUT & 5\% & 0.0112 & 0.0555 & 0.1237 & 16.99 & 11.76 \\
\bottomrule
    \end{autowidthtabular}
\end{table}

\subsection{Results by Task Type}
\label{app:task_type_results}

Table~\ref{tab:category_details} reports every category in the native
\emph{problem type} field of each development benchmark, with categories
listed alphabetically. \(N\) is the number of QA examples in the category,
unchanged across the five evaluation budgets. The final three columns report
arithmetic averages of metrics computed separately at each budget.
Delta is the adapted-minus-Unadapted accuracy change in percentage points.
Repair rate is the percentage of initially incorrect predictions that become
correct, and Regress rate is the percentage of initially correct predictions
that become incorrect.

DailyOmni and AVUT show positive average accuracy gains in every listed task
category. WorldSense exhibits more varied changes, with larger gains in
attribute reasoning, event recognition, and video-emotion tasks, alongside
unchanged or lower accuracy in some other categories. These results show
that adaptation benefits a range of tasks, while the magnitude of improvement
varies by task type.

\begingroup
\appendixlongtablestyle
\begin{longtable}{llrrrr}
\caption{Five-budget average behavior by benchmark-native problem type.
The model is trained with DivPrune at 5\% nominal retention and evaluated
with DivPrune across five deployment budgets.}
\label{tab:category_details}\\
\toprule
Benchmark & Problem type & \(N\) & Delta (pp) & Repair (\%) & Regress (\%) \\
\midrule
\endfirsthead
\multicolumn{6}{c}{\tablename\ \thetable\ continued}\\
\toprule
Benchmark & Problem type & \(N\) & Delta (pp) & Repair (\%) & Regress (\%) \\
\midrule
\endhead
\midrule
\multicolumn{6}{r}{Continued on next page}\\
\endfoot
\bottomrule
\endlastfoot
WorldSense & Action Counting & 165 & 0.24 & 6.00 & 13.57 \\
WorldSense & Anomaly Recognition & 79 & 1.27 & 3.71 & 2.58 \\
WorldSense & Attribute Reasoning & 121 & 3.97 & 8.32 & 1.11 \\
WorldSense & Attribute Recognition & 181 & 0.00 & 4.10 & 4.62 \\
WorldSense & Audio Change & 83 & 1.45 & 6.43 & 6.90 \\
WorldSense & Audio Counting & 90 & 1.11 & 7.19 & 9.73 \\
WorldSense & Audio Recognition & 116 & 0.00 & 3.62 & 3.25 \\
WorldSense & Audio Source Localization & 120 & 1.17 & 4.88 & 4.00 \\
WorldSense & Causal Reasoning & 151 & 0.00 & 3.39 & 2.90 \\
WorldSense & Emotion Change & 96 & 3.54 & 9.89 & 3.54 \\
WorldSense & Event Recognition & 122 & 3.93 & 9.51 & 1.70 \\
WorldSense & Event Sorting & 171 & 1.64 & 6.23 & 3.96 \\
WorldSense & Hallucination & 90 & -4.00 & 3.46 & 11.94 \\
WorldSense & Human Emotions & 98 & 2.65 & 7.79 & 1.98 \\
WorldSense & Human Interaction & 132 & 0.30 & 4.99 & 6.77 \\
WorldSense & Human-object Interaction & 127 & -0.63 & 4.44 & 6.30 \\
WorldSense & Object Counting & 205 & 2.24 & 6.61 & 8.47 \\
WorldSense & Object Existence Recognition & 107 & 2.99 & 8.54 & 4.08 \\
WorldSense & Object State Change & 95 & 1.05 & 6.78 & 6.01 \\
WorldSense & Relation Reasoning & 84 & 1.67 & 5.15 & 2.63 \\
WorldSense & Scene Recognition & 81 & 1.98 & 6.60 & 3.22 \\
WorldSense & Spatial Relation & 197 & 0.61 & 3.65 & 7.23 \\
WorldSense & Temporal Localization & 169 & 0.00 & 6.19 & 11.70 \\
WorldSense & Temporal Prediction & 110 & 0.91 & 5.58 & 4.35 \\
WorldSense & Text and Diagram Understanding & 134 & 1.19 & 6.60 & 4.82 \\
WorldSense & Video Emotions & 48 & 4.17 & 17.95 & 1.78 \\
\addlinespace[2pt]
DailyOmni & AV Event Alignment & 238 & 0.34 & 6.02 & 7.69 \\
DailyOmni & Comparative & 131 & 2.29 & 10.42 & 1.65 \\
DailyOmni & Context understanding & 193 & 3.01 & 9.33 & 4.91 \\
DailyOmni & Event Sequence & 306 & 2.75 & 8.71 & 4.09 \\
DailyOmni & Inference & 154 & 3.12 & 15.25 & 1.43 \\
DailyOmni & Reasoning & 175 & 2.74 & 9.84 & 0.00 \\
\addlinespace[2pt]
AVUT & Audio Character Matching & 417 & 2.35 & 11.06 & 4.23 \\
AVUT & Audio Content Counting & 118 & 3.73 & 6.76 & 1.95 \\
AVUT & Audio Event Location & 170 & 0.24 & 7.72 & 14.03 \\
AVUT & Audio Information Extraction & 326 & 0.86 & 11.16 & 1.59 \\
AVUT & Audio OCR Matching & 311 & 0.51 & 7.50 & 3.45 \\
AVUT & Audio Object Matching & 392 & 1.22 & 5.77 & 3.00 \\
 
\end{longtable}
\endgroup

\endgroup

\end{document}